\documentclass[preprint,10pt,authoryear]{elsarticle}

\usepackage{graphicx}
\usepackage{url}
\usepackage{enumitem}
\usepackage{amsfonts}
\usepackage{amsmath}
\usepackage{amssymb}
\usepackage{booktabs}
\usepackage{enumitem}
\usepackage{boldline}
\usepackage{threeparttable}
\usepackage{array}
\usepackage{float}
\usepackage{textcomp}
\usepackage{mdwmath}
\usepackage{mdwtab}
\usepackage{subfig}
\usepackage{amsmath}
\usepackage{color}
\usepackage{tikz}
\usepackage{multirow}
\usepackage{bbding}
\usepackage{caption}
\usepackage{bm}
\usepackage{amsthm}
\usepackage{dblfloatfix}
\usepackage{hyperref}
\usepackage{algorithm,algpseudocode}
\usepackage{lipsum}
\usepackage{etoolbox}
\usepackage{lineno}
\usepackage{tikz}
\usepackage{subcaption}
\usepackage[margin=2cm]{geometry}

\definecolor{navy}{RGB}{0, 0, 128}



\algnewcommand\algorithmicinput{\textbf{Input:}}
\algnewcommand\Input{\item[\algorithmicinput]}
\algnewcommand\algorithmicoutput{\textbf{Output:}}
\algnewcommand\Output{\item[\algorithmicoutput]}
\algnewcommand\algorithmicParameter{\textbf{Parameters:}}
\algnewcommand\Parameters{\item[\algorithmicParameter]}

\journal{Multimodal Transportation}

\biboptions{authoryear}

\begin{document}

\begin{frontmatter}

% figure caption style
\captionsetup[figure]{labelfont={bf},labelformat={default},labelsep=period,name={Fig.}}
% table caption style
\captionsetup[table]{labelfont={bf},labelformat={default},labelsep=period,name={Table}}

\title{Spatio-temporal dual-stage hypergraph MARL for human-centric multimodal corridor traffic signal control}

%\author{Xiaocai Zhang\fnref{myfootnote}}

\author[mymainaddress]{Xiaocai Zhang\corref{mycorrespondingauthor}}
\ead{xiaocai.zhang@unimelb.edu.au}

\author[mymainaddress]{Neema Nassir}
\ead{neema.nassir@unimelb.edu.au}

\author[mymainaddress]{Milad Haghani}
\ead{milad.haghani@unimelb.edu.au}

\cortext[mycorrespondingauthor]{Corresponding author}

\address[mymainaddress]{Department of Infrastructure Engineering, Faculty of Engineering and Information Technology, The University of Melbourne, VIC 3010, Australia}
% \address[mysecondaryaddress]{Address 2}

\begin{abstract}
Human-centric traffic signal control in corridor networks must increasingly account for multimodal travelers, particularly high-occupancy public transportation, rather than focusing solely on vehicle-centric performance. This paper proposes STDSH-MARL (Spatio-Temporal Dual-Stage Hypergraph based Multi-Agent Reinforcement Learning), a multi-agent deep reinforcement learning framework that follows a centralized training and decentralized execution paradigm. The proposed method captures spatio-temporal dependencies through a novel dual-stage hypergraph attention mechanism that models interactions across both spatial and temporal hyperedges. In addition, a hybrid discrete action space is introduced to jointly determine the next signal phase configuration and its corresponding green duration, enabling more adaptive signal timing decisions. Experiments conducted on a corridor network under five traffic scenarios demonstrate that STDSH-MARL achieves strong overall multimodal performance, with substantial and relatively consistent reductions in tram waiting time, while improvements in bus waiting time are more variable across traffic scenarios. These results highlight the trade-off among overall network efficiency, tram priority, and bus service quality. Compared with state-of-the-art baseline methods, the proposed approach achieves superior overall performance. Further ablation studies confirm the contribution of each component of STDSH-MARL, with temporal hyperedges identified as the most influential factor driving the observed performance gains.

\end{abstract}

\begin{keyword}
Traffic signal control \sep hypergraph \sep human-centric \sep multi-agent \sep deep reinforcement learning.
\end{keyword}

\end{frontmatter}

\section{Introduction}
Traffic signals coordinate movements at intersections to enhance the efficiency of transportation networks. Effective traffic signal control remains a challenging but active research problem within transportation due to its complex, dynamic, and stochastic characteristics \citep{wei2019presslight,zhang2026human}. Traditional fixed-time Traffic Signal Control (TSC) algorithms become inefficient in managing congestion within highly dynamic traffic scenarios. Although manual traffic control can be effective in specific situations, it often leads to inefficient use of both time and manpower. Early-stage adaptive TSC methods tackle optimization challenges to develop efficient coordination and control strategies. Notable TSC solutions, including Split Cycle Offset Optimisation Technique (SCOOT) \citep{hunt1982scoot} and Sydney Coordinated Adaptive Traffic System (SCATS) \citep{stevanovic2009scoot}, have been implemented in many cities around the world \citep{chu2019multi}. However, these adaptive TSC systems, like SCATS, mainly address vehicle traffic and often require further enhancements to better accommodate other modes, such as public transportation, pedestrians, and cyclists. To serve integrated multimodal transportation systems more effectively, SCATS and similar adaptive approaches require more innovative adaptations \citep{mccannfeet}.

In order to elevate the efficiency of TSC, the paradigm of Deep Reinforcement Learning (DRL) has been extensively leveraged to handle the complex and fluctuating environmental dynamics in traffic and transportation systems \citep{chan2026multi}. DRL is particularly effective in TSC due to its ability to autonomously discover optimal strategies through a process of trial and error, effectively adapt to evolving conditions, and make informed decisions grounded in intricate and variable traffic patterns. This approach has been shown to surpass conventional transportation approaches in managing traffic flows, as evidenced by several recent studies \citep{yazdani2023intelligent,zhang2025towards,Chan2026regulating}. These studies highlight the substantial benefits of DRL, showcasing its potential to revolutionize traffic management by enhancing decision-making processes and response times to real-time traffic situations.

In the early stages, the applications of DRL in TSC were primarily concentrated on individual intersections. Subsequently, techniques like Q-learning \citep{prabuchandran2014multi}, DQN \citep{ge2019cooperative}, DDDQN \citep{gong2019decentralized}, and hybrid methods that integrate fuzzy control with DRL \citep{kumar2020fuzzy} were gradually extended to optimize the signal control of multiple intersections. Coordinating traffic signals effectively is essential to ensure that all traffic lights within a network collaborate to enhance overall traffic flow efficiency. This paper aims to develop advanced DRL methodologies for TSC at multiple intersections within the same corridor. However, most prior DRL-based multi-intersection coordination studies primarily optimize vehicle-centric objectives, such as maximizing throughput \citep{duan2025bayesian} and minimizing waiting time \citep{xu2024graph} or congestion. Yet such vehicle-centric metrics can overlook how delays are distributed across modes and travelers, especially when public transport carries far more passengers per vehicle. This motivates the exploration of more humanized and human-centric objectives, such as number of passengers experiencing delay, particularly for high-occupancy public transportation, to maintain equity and human-centric utilitarianism among different modalities of transportation, within DRL frameworks for traffic and social management.

In this paper, we introduce a novel DRL framework tailored for multi-agent, multimodal TSC across networks of multiple intersections. The proposed framework follows an actor–critic architecture under a Centralized Training and Decentralized Execution (CTDE) paradigm, where centralized critics leverage global network-level information during training, while decentralized actors operate solely on local observations at execution time. To effectively capture complex spatial and temporal dependencies among intersections, we design a dual-stage hypergraph (DSH) representation that models both spatial and temporal interactions. Building on this design, we term our approach STDSH-MARL (\textbf{S}patio-\textbf{T}emporal \textbf{D}ual-\textbf{S}tage \textbf{H}ypergraph-based \textbf{M}ulti-\textbf{A}gent \textbf{R}einforcement \textbf{L}earning). The transportation modalities involved include vehicles and public transportation of buses and trams to accommodate a large number of passengers.
In this study, the human-centric objective is defined at the passenger level rather than the vehicle level. Each traveler contributes equally to the reward, regardless of whether they are in a private vehicle, bus, or tram. As a result, high-occupancy public transport may receive implicit operational preference because delaying one bus or tram affects more travelers, not because the reward assigns explicit mode-specific priority weights.

\begin{figure}[htbp]
\begin{center}
\includegraphics[width=0.999\textwidth]{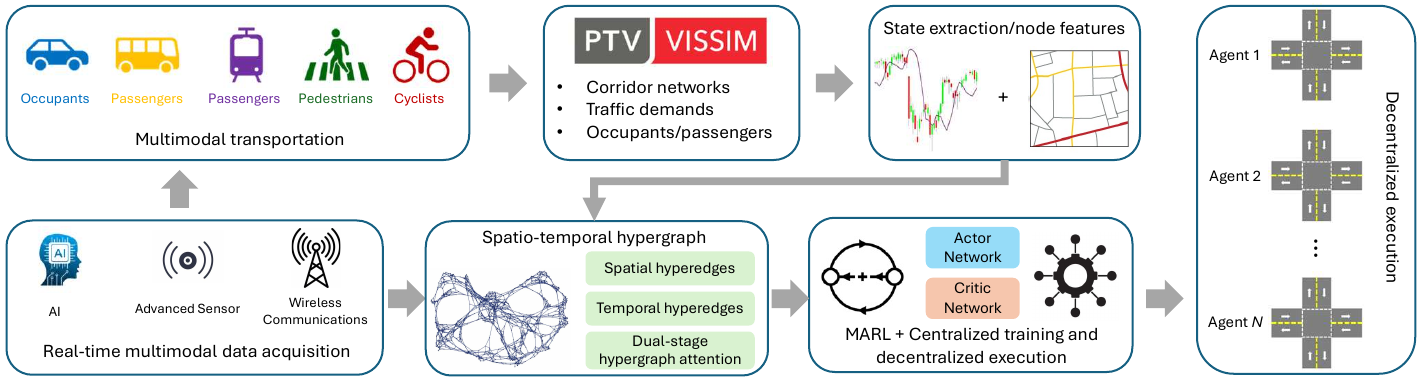}
\caption{Overview of the proposed STDSH-MARL framework for multimodal corridor traffic signal control.}
\label{tsc_source}
\end{center}
\end{figure}

As shown in Figure \ref{tsc_source}, the proposed framework considers transportation users across multiple modes, including vehicle occupants, bus and tram passengers, pedestrians, and cyclists. Such multimodal user information is essential for human-centric traffic signal optimization, as it enables the control objective to move beyond vehicle-based efficiency and account for the mobility experience of different user groups. Recent advances in Artificial Intelligence (AI), sensing technologies, wireless communications, and connected infrastructure have significantly improved the feasibility of real-time multimodal data acquisition. For example, vehicle occupancy can be estimated using vision-based or learning-based methods \citep{jiang2023pa,papakis2021convolutional}, pedestrian volumes can be inferred from Bluetooth Low Energy scanners \citep{gong2022using}, and on-board passenger counts for public transport can be predicted using advanced machine learning models \citep{roncoli2023estimating}. These multimodal data streams can then be integrated with traffic demands and network information in PTV VISSIM to extract state and node features. Based on these features, a spatio-temporal hypergraph is constructed to capture spatial and temporal dependencies, which supports centralized training and decentralized execution of multi-agent traffic signal control.

Unlike prior spatio-temporal graph-based traffic signal control methods that mainly rely on pairwise adjacency modeling or separate spatial and temporal modules, the proposed method represents the corridor as a unified spatio-temporal hypergraph. In this formulation, spatial hyperedges capture cross-intersection dependencies at each time step, while temporal hyperedges capture historical dependencies of each intersection across time. On top of this representation, we introduce a dual-stage hypergraph attention mechanism that first learns the importance of nodes within each hyperedge and then learns the importance of different hyperedges for each node. This enables higher-order node–hyperedge–node reasoning beyond conventional graph aggregation. Furthermore, this hypergraph representation is integrated into the centralized critic under the CTDE framework, while the decentralized actors make decisions using local observations during execution.

The main \textbf{contributions} of this study are summarized as follows:
\begin{enumerate}
\item We introduce a passenger-based human-centric objective for multimodal traffic signal control, in which each traveler is counted equally across modes;
\item We include multimodal transportation optimization by considering the priority for high-capacity public transit (e.g., trams) to move more passengers efficiently;
\item We design an adaptive hybrid action space including comprehensive signal configurations as well as their corresponding green timing;
\item We present a multi-agent RL framework employing a novel spatio-temporal dual hypergraph attention mechanism in modeling corridor network;
\item We have conducted comprehensive testing across five typical traffic scenarios to validate the efficiency and robustness of our proposed DRL model in multi-agent environments.
\end{enumerate}

The structure of subsequent sections of this paper is organized as follows: Section \ref{relatedwork} provides an in-depth review of existing research on multi-agent DRL approaches for TSC. Section \ref{method} details the proposed methodology, including the spatio-temporal hypergraph, dual-stage hypergraph attention, state representation, action space, reward mechanism, and MARL training process. Section \ref{experiment} discusses the experimental results, analyses, and ablation studies. Finally, Section \ref{conclusion} concludes the study and outlines directions for future research.

\section{Related work}
\label{relatedwork}
We conducted an extensive literature review of 45 recent peer-reviewed papers on multi-agent DRL in TSC across multiple intersections. These papers were published in prestigious journals and conferences, including Transportation Research Part C: Emerging Technologies, IEEE Transactions on Intelligent Transportation Systems, Neurocomputing, the ACM SIGKDD Conference on Knowledge Discovery and Data Mining, and the International Joint Conference on Artificial Intelligence (IJCAI). In the following sections, we review existing TSC methods that utilize DRL, examining them from four perspectives: traffic simulation environment, reward mechanism, graph-based network representation, and multi-agent DRL framework.

\subsection{Traffic simulation environment}
Microscopic traffic simulation is widely used to replicate realistic traffic environments for multi-agent traffic signal control research based on DRL. Five popular simulation tools are commonly employed in this field: SUMO\footnote[1]{\url{https://eclipse.dev/sumo/}}, CityFlow\footnote[2]{\url{https://cityflow-project.github.io/}}, VISSIM\footnote[3]{\url{https://www.ptvgroup.com/en/products/ptv-vissim}}, Aimsun\footnote[4]{\url{https://www.aimsun.com/}}, and Paramics\footnote[5]{\url{https://www.systra.com/digital/solutions/transport-planning/paramics/}}. Table \ref{table_simulation} lists the traffic simulation tools used by each study. The majority of existing studies, around 57.78\% (26 out of 45), use SUMO for environment simulation, probably due to its free availability, which is advantageous for budget-limited projects. CityFlow ranks as the second most popular choice, utilized in 10 studies. PTV VISSIM is third, used in 5 studies, Aimsun has been applied in 2 studies for multi-agent traffic signal control. While Paramics has been used in only one study. SUMO and CityFlow are freely available, while VISSIM, Aimsun, and Paramics are commercial software, which can be costly. While SUMO is functional, its visualization is less advanced compared to commercial software. CityFlow is specifically designed for reinforcement learning applications in traffic signal control, making it less versatile than other simulators. Paramics offers detailed microscopic simulation and real-time traffic management capabilities, making it particularly suitable for testing adaptive signal control and ITS applications. VISSIM excels in modeling complex vehicle interactions and mixed traffic, whereas Aimsun supports both microscopic and mesoscopic simulations and can handle large networks. Additionally, VISSIM, Aimsun, and Paramics can be integrated with other software and databases, enhancing their functionality.

\begin{table}[htbp]
\centering
\caption{Summary of traffic simulation environments}
\label{table_simulation}
\begin{tabular}{p{3.0cm}p{10.5cm}p{1.5cm}}
\hline
Traffic Simulation & References & Quantity \\
\hline
SUMO & \cite{wang2021adaptive,li2021network,su2023emvlight,song2024cooperative,wang2024large,bie2024multi,yang2023hierarchical,yang2021ihg,ren2024two,xu2024graph,zhang2022distributed,hu2024multi,tan2020multi,chu2019multi,wu2020multi1,bokade2023multi,ge2021multi,luo2024reinforcement,liu2023traffic,nie2025cmrm,jia2025multi,fereidooni2025multi,shen2025hierarchical,chu2016large,ma2020feudal,tao2023network} & 26\\
\hline
CityFlow & \cite{zhu2023multi,liu2023multiple,wu2022distributed,liu2023gplight,zhang2022neighborhood,wang2020stmarl,liu2025globallight,lai2025llmlight,satheesh2025constrained,wang2025towards} & 10\\
\hline
VISSIM & \cite{jiang2021distributed,wang2021gan,wu2020multi2,lee2019reinforcement,zhang2025towards} & 5\\
\hline
Aimsun & \cite{abdoos2020cooperative,yoon2025decentralized} & 2\\
\hline
Paramics & \cite{el2013multiagent} & 1\\
\hline
Not mentioned & \cite{wang2020large} & 1\\
\hline
\end{tabular}
\end{table}

\subsection{Reward mechanism}
The reward mechanism is a key component in developing DRL models, as it guides the learning process of an agent by providing feedback during interactions with its environment. Most of the studies reviewed primarily define rewards from the perspective of the vehicle, with only 2 studies out of 45 also considering multimodal transport, like pedestrians, in the reward function design.
For example, \cite{wu2020multi1} developed a comprehensive reward function based on a weighted value incorporating six factors: total vehicle queue length, cumulative vehicle waiting time, vehicle delay, total vehicle throughput, traffic light status, and pedestrian waiting times at intersections. \cite{zhang2025towards} developed a fairness-based traffic signal control method that uses travelers affected by delay as the reward function, which includes travelers from private vehicles, public transportation and pedestrians.
In studies focused solely on vehicles, reward functions are typically based on factors like vehicle delay, queue length (or the number of vehicles waiting at intersections), waiting time, throughput, and speed, either individually or in combination. Table \ref{table_reward_1} and \ref{table_reward_2} provide a detailed summary of reward definitions across studies. Queue length is a commonly used metric, with 13 studies relying exclusively on it as an indicator, as shown in the first row of Table \ref{table_reward_1}.
Additionally, some studies combine queue length with other indicators. For example, 5 studies use both queue length and vehicle waiting time, making this the second most popular reward mechanism. \cite{su2023emvlight} introduced the concept of 'pressure', defined as the difference between traffic density on incoming and outgoing lanes. Other research works used changes between consecutive time steps as rewards; for instance, \cite{jiang2021distributed} and \cite{luo2024reinforcement} defined rewards as the change in vehicle queue length between timesteps, capturing evolving dynamics during learning. Similarly, \cite{li2021network} leveraged the change in vehicle delay between consecutive times as the reward function.
The most comprehensive reward structure was constructed by \cite{hu2024multi}, incorporating six indicators: vehicle throughput, queue length, phase switch penalty, minor street phase penalty, lane delay, and waiting time. Also, \cite{shen2025hierarchical} constructed a reward based on six indicators: total delay, queue length, waiting time, throughput, average travel time and average speed.

\subsection{Graph-based network representation}
Graph-based representations model the traffic network as a graph, where intersections are treated as nodes and roads as edges, allowing the use of Graph Neural Networks (GNNs), Convolutional Neural Networks (CNNs), and Dynamic Spatio-Temporal Hypergraph (DSTH) to capture complex structural and relational dependencies within the network. These models are especially suited for large-scale traffic control systems, where understanding interconnections between intersections is crucial for effective coordination.
A wide range of graph-based techniques have been applied. Common GNN variants include Graph Convolutional Networks (GCNs), Graph Attention Networks (GATs), Graph Sample and Aggregate Network (GraphSAGE), and hierarchical or spatial-attention models such as Stochastic Aggregation GNN (SA-GNN), Heterogeneous Graph Attention Network (HGAT), and Heterogeneous GCN (HGCN). These approaches are capable of extracting high-level representations that reflect both topological structure and dynamic traffic conditions.
For instance, \cite{wang2024large} employed GATs with multi-head attention to aggregate local information within intersections.
\cite{yoon2025decentralized} modeled the road network as a bi-directional link graph and applied a message-passing GNN that iteratively updates edge and node embeddings via multi-layer perceptron with permutation-invariant aggregation to learn traffic dynamics.
\cite{chu2016large} represented the traffic network as a stack of 2D images, with each pixel corresponding to a spatial location (e.g., traffic light or intersection), and image channels encoding different types of traffic-related data. CNNs were then used to extract spatial features from these images.
Additionally, graph decomposition techniques such as Laplacian matrices and the Normalized Cut (Ncut) algorithm facilitate structural learning by segmenting the network into regions based on connectivity and traffic flow. In the study by \cite{ma2020feudal}, the network is partitioned into disjoint subgraphs, and a hierarchical structure is built to enhance global coordination. Similarly, \cite{tao2023network} applied the Ncut method to cluster intersections into spatially compact regions with similar congestion profiles.
\cite{wang2025towards} extended this concept by introducing a DSTH, where hyperedges connect multiple intersections instead of just forming pairwise relationships. These connections are determined by dynamically computed reconstruction coefficients. 
\cite{nie2025cmrm} introduced a Cooperation Enhancement Module built on a GAT network, which lets each intersection dynamically aggregate neighbouring agents’ state information, so agents can better capture the complex, time-varying interactions across the network.

\begin{table}[htbp]
\footnotesize
\centering
\caption{Summary of factors considered in reward mechanism for each reference (\romannumeral1)}
\label{table_reward_1}
\begin{threeparttable}
\begin{tabular}{p{6.0cm}p{8.3cm}p{1.5cm}}
\hline
Reward Components & References & Quantity \\
\hline
Queue length (veh) & \cite{zhu2023multi,liu2023multiple,wang2020large,liu2023gplight,tan2020multi,bokade2023multi,zhang2022neighborhood,wang2020stmarl,fereidooni2025multi,liu2025globallight,lai2025llmlight,chu2016large,wang2025towards} & 13 \\
\hline
Waiting time (veh)& \cite{wang2024large,yang2023hierarchical,yang2021ihg}; & \multirow{2}{*}{5} \\
Queue length (veh) & \cite{chu2019multi,liu2023traffic} & \multirow{2}{*}{} \\
\hline
Difference of queue lengths between two consecutive timesteps (veh) & \cite{jiang2021distributed,luo2024reinforcement} & 2 \\
\hline
Waiting time (veh) & \cite{wu2022distributed,xu2024graph} & 2 \\
\hline
Throughput (veh) & \multirow{6}{*}{\cite{hu2024multi}} & \multirow{6}{*}{1} \\
Queue length (veh) & \multirow{6}{*}{} & \multirow{6}{*}{} \\
Phase switch penalty & \multirow{6}{*}{} & \multirow{6}{*}{} \\
Minor street phase penalty & \multirow{6}{*}{} & \multirow{6}{*}{} \\
Lane delay/speed loss (veh) & \multirow{6}{*}{} & \multirow{6}{*}{} \\
Waiting time (veh) & \multirow{6}{*}{} & \multirow{6}{*}{} \\
\hline
Total delay (veh) & \multirow{6}{*}{\cite{shen2025hierarchical}} & \multirow{6}{*}{1} \\
Queue length (veh) & \multirow{6}{*}{} & \multirow{6}{*}{} \\
Waiting time (veh) & \multirow{6}{*}{} & \multirow{6}{*}{} \\
Throughput (veh) & \multirow{6}{*}{} & \multirow{6}{*}{} \\
Average travel time (veh) & \multirow{6}{*}{} & \multirow{6}{*}{} \\
Average speed (veh) & \multirow{6}{*}{} & \multirow{6}{*}{} \\
\hline
Waiting time (veh) & \multirow{5}{*}{\cite{wu2020multi1}} & \multirow{5}{*}{1} \\
Waiting time (ped) & \multirow{5}{*}{} & \multirow{5}{*}{} \\
Queue length (veh) & \multirow{5}{*}{} & \multirow{5}{*}{} \\
Throughput (veh) & \multirow{5}{*}{} & \multirow{5}{*}{} \\
Traffic light blinking condition & \multirow{5}{*}{} & \multirow{5}{*}{} \\
\hline
Waiting time (veh) & \multirow{4}{*}{\cite{ge2021multi}} & \multirow{4}{*}{1} \\
Queue length (veh) & \multirow{4}{*}{} & \multirow{4}{*}{} \\
Total number of vehicles & \multirow{4}{*}{} & \multirow{4}{*}{} \\
Speed (veh) & \multirow{4}{*}{} & \multirow{4}{*}{} \\
\hline
Number of approaching vehicles & \multirow{4}{*}{\cite{ma2020feudal}} & \multirow{4}{*}{1} \\
Cumulative waiting time (veh) & \multirow{4}{*}{} & \multirow{4}{*}{} \\
Number of vehicles reaching destination & \multirow{4}{*}{} & \multirow{4}{*}{} \\
Traffic flow liquidity & \multirow{4}{*}{} & \multirow{4}{*}{} \\
\hline
Traffic density (veh) & \multirow{3}{*}{\cite{wang2021adaptive}} & \multirow{3}{*}{1} \\
Speed (veh) & \multirow{3}{*}{} & \multirow{3}{*}{} \\
Distance between intersections & \multirow{3}{*}{} & \multirow{3}{*}{} \\
\hline
\end{tabular}
        \textit{Notes:} “veh” means vehicle and “ped” stands for pedestrian.
\end{threeparttable}
\end{table}

\begin{table}[htbp]
\footnotesize
\centering
\caption{Summary of factors considering for reward mechanism for each reference (\romannumeral2)}
\label{table_reward_2}
\begin{threeparttable}
\begin{tabular}{p{6.0cm}p{8.3cm}p{1.5cm}}
\hline
Reward Components & References & Quantity \\
\hline
Travel time delay (veh) & \multirow{3}{*}{\cite{song2024cooperative}} & \multirow{3}{*}{1} \\
Queue length (veh) & \multirow{3}{*}{} & \multirow{3}{*}{} \\
Throughput (veh) & \multirow{3}{*}{} & \multirow{3}{*}{} \\
\hline
Waiting time (veh) & \multirow{3}{*}{\cite{bie2024multi}} & \multirow{3}{*}{1} \\
Queue length (veh) & \multirow{3}{*}{} & \multirow{3}{*}{} \\
Throughput (veh) & \multirow{3}{*}{} & \multirow{3}{*}{} \\
\hline
Queue length (veh) & \multirow{3}{*}{\cite{zhang2022distributed}} & \multirow{3}{*}{1} \\
Traffic density (veh) & \multirow{3}{*}{} & \multirow{3}{*}{} \\
Cumulative stop delay (veh) & \multirow{3}{*}{} & \multirow{3}{*}{} \\
\hline
Waiting time (veh) & \multirow{3}{*}{\cite{nie2025cmrm}} & \multirow{3}{*}{1} \\
Emission (veh) & \multirow{3}{*}{} & \multirow{3}{*}{} \\
Cumulative collision risk index (veh) & \multirow{3}{*}{} & \multirow{3}{*}{} \\
\hline
Waiting time (veh) & \multirow{3}{*}{\cite{jia2025multi}} & \multirow{3}{*}{1} \\
Throughput (veh) & \multirow{3}{*}{} & \multirow{3}{*}{} \\
Variance of green time across phases & \multirow{3}{*}{} & \multirow{3}{*}{} \\
\hline
Total number of vehicles & \multirow{2}{*}{\cite{lee2019reinforcement}} & \multirow{2}{*}{1} \\
Speed (veh) & \multirow{2}{*}{} & \multirow{2}{*}{} \\
\hline
Queue length (veh) & \multirow{2}{*}{\cite{yoon2025decentralized}} & \multirow{2}{*}{1} \\
Throughput (veh) & \multirow{2}{*}{} & \multirow{2}{*}{} \\
\hline
Waiting time (veh) & \multirow{2}{*}{\cite{ren2024two}} & \multirow{2}{*}{1} \\
Emission (veh) & \multirow{2}{*}{} & \multirow{2}{*}{} \\
\hline
Queue length (veh) & \multirow{2}{*}{\cite{tao2023network}} & \multirow{2}{*}{1} \\
Average speed (veh) & \multirow{2}{*}{} & \multirow{2}{*}{} \\
\hline
Number of moving vehicles & \multirow{2}{*}{\cite{satheesh2025constrained}} & \multirow{2}{*}{1} \\
Number of waiting vehicles & \multirow{2}{*}{} & \multirow{2}{*}{} \\
\hline
Difference of delays between two consecutive timesteps (veh) & \cite{li2021network} & 1 \\
\hline
Average delay time (veh) & \cite{abdoos2020cooperative} & 1 \\
\hline
Cumulative delay (veh) & \cite{el2013multiagent} & 1 \\
\hline
Pressure (veh) & \cite{su2023emvlight} & 1 \\
\hline
Throughput (veh) & \cite{wu2020multi2} & 1 \\
\hline
Speed (veh) & \cite{wang2021gan} & 1 \\
\hline
Number of delayed travelers & \cite{zhang2025towards} & 1 \\
\hline
\end{tabular}
        \textit{Notes:} “veh” means vehicle and “ped” stands for pedestrian.
\end{threeparttable}
\end{table}

\subsection{Multi-agent DRL framework}
The fundamental DRL framework for multi-agent TSC can be categorized into three types based on the learning process: value-based DRL methods, policy-based DRL methods, and hybrid DRL methods. Each of these will be explained in the following sections.

\subsubsection{Value-based DRL methods}
Value-based DRL focuses on learning the value of each action in a given state without explicitly learning the policy. The policy is derived by selecting actions that maximize the estimated values \citep{mckenzie2022modern}. For multi-agent TSC, value-based DRL methods primarily include Q-learning and deep Q-learning variants, such as Deep Q-Network (DQN), Double Deep Q-Network (DDQN), Double Dueling Deep Q-Learning Network (DDDQN), and QMIX. Table \ref{value_based} summarizes the value-based DRL approaches from the reviewed papers.

Traditional Q-learning methods have been employed in 4 studies. For example, \cite{zhu2023multi} proposed a multi-agent DRL approach using broad learning architectures \citep{gong2021research}, which significantly reduces model complexity and, consequently, training time, without sacrificing performance. Q-learning was utilized within this broader framework.
In contrast, \cite{abdoos2020cooperative} introduced an innovative TSC method that integrates game theory with DRL. This method features a two-mode agent architecture in which each intersection is controlled by an agent that operates independently during normal traffic conditions and cooperatively when traffic becomes congested. Each agent utilizes Q-learning to manage its intersection under normal conditions. However, in times of congestion, the agents shift to a cooperative mode, using game theory to dynamically coordinate traffic signals with adjacent agents.
\cite{el2013multiagent} proposed MARLIN-ATSC, a decentralized multi-agent RL system for network-wide adaptive signal control that coordinates neighboring intersections via modular Q-learning and learned neighbor-policy models.

Multi-agent DRL based on DQN is the most popular method, employed by a total of 15 studies.
\cite{wang2021adaptive} proposed a cooperative, group-based multi-agent DRL framework that uses k-Nearest-Neighbor (kNN) for state representation to enhance coordination among agents. They adapted the DQN framework to a multi-agent environment for Q-value approximation.
\cite{wang2024large} introduced a multi-agent DRL approach tailored for complex, large-scale traffic networks, where each intersection is modeled as a graph with lanes acting as nodes and traffic relationships forming the edges. Graph Attention Network (GAT) is utilized to learn lane embeddings that encapsulate spatial and geometric features of intersections. These embeddings are then fed into DQN for the evaluation of Q-values.
\cite{bie2024multi} presented a multi-agent DRL framework that tackles intersection heterogeneity by employing Graph Attention Networks (GATs) to capture interactions and dependencies between intersections. The core element of this DRL approach is DQN, which aims to select optimal actions by maximizing Q-values.
\cite{liu2023multiple} employed a decentralized form of DQN, in which each agent operates independently but shares some information with neighboring agents.
\cite{xu2024graph} addressed the issue of missing data in TSC by introducing a Wasserstein Generative Adversarial Network (WGAN) to estimate the state space and maintain data integrity. They also utilized a Graph Neural Network (GNN) to aggregate state features from multiple agents, which is crucial for informed decision-making. Each agent then applied decentralized DQN-based learning to select the optimal action.
In the study by \cite{wang2021gan}, a Generative Adversarial Network (GAN) framework was employed for traffic data recovery, utilizing a decentralized model based on DQN.
\cite{tan2020multi} utilized the bootstrapped DQN algorithm, which enhances exploration through an ensemble of behavior policies. This approach demonstrated superior efficiency and robustness compared to standard DQN by managing uncertainty more effectively.
\cite{wu2020multi2} demonstrated a decentralized multi-agent DRL approach based on DQN, applying transfer learning to assess the robustness of the algorithm and traditional control methods under varying traffic scenarios, such as changes in traffic volume, flow patterns, and sensor reliability.
\cite{zhang2022neighborhood} implemented a decentralized multi-agent DRL where each agent learns independently from its own observations as well as those from neighboring intersections, utilizing Hysteretic DQN for strategy optimization. Hysteretic DQN, which is a variant of the standard DQN, introduces a novel update mechanism designed to address the complexities in multi-agent environments.
\cite{liu2023traffic} introduced a multi-agent DRL strategy for optimizing traffic light control in areas affected by epidemics, incorporating a Convolutional Neural Network (CNN) to process state representations within the DQN framework. Meanwhile, \cite{wang2020stmarl} discussed a spatio-temporal multi-agent DRL framework that integrates GNN, RNN, and DQN. RNNs are employed to capture the temporal dynamics of traffic, while GNNs model cooperative structures among traffic lights based on an adjacency graph, enabling each traffic light to make decisions through deep Q-learning.

\cite{wang2020large} addressed the issue of large-scale traffic signal control with a scalable independent double DQN (DDQN) method aimed at reducing the overestimation issues found in standard DQN algorithms. The approach incorporates an upper confidence bound policy to balance the exploration-exploitation trade-off.
\cite{liu2023gplight} created a grouped multi-agent DRL framework for managing large-scale traffic signal control by clustering agents (intersections) based on environmental similarity. A Graph Convolutional Network (GCN) extracts features from the traffic network, modeled as a graph, and QMIX is implemented for policy learning. Agents within the same group share a neural network model and parameters, facilitating efficient learning and decision-making.
\cite{bokade2023multi} demonstrated a centralized training with decentralized execution approach for inter-agent communication in large-scale traffic control. Using QMIX, agents learn to identify which parts of their messages need to be shared with others to achieve effective coordination.
\cite{liu2025globallight} proposed a method that (i) uses a multi-head GAT to extract local multi-dimensional features, (ii) mines global similarities via two representation-space losses to group distant yet similar intersections and share policy parameters, and (iii) integrates these in a QMIX learner.
\cite{hu2024multi} modeled traffic signal coordination using a multi-agent system that incorporates both local and global agents with a DDQN approach. The system comprises seven local agents, one for each intersection, which operate independently but are coordinated by a global agent. This setup allows for localized decision-making while aligning with global objectives.
\cite{jiang2021distributed} proposed a distributed multi-agent DRL approach that utilizes graph decomposition to handle large-scale traffic signal control. The network of intersections is segmented into subgraphs according to connectivity, with each controlled by a collective of agents operating within the DRL framework. They employ the framework of DDDQN to enhance learning efficiency. DDDQN combines the advantages of DDQN to minimize Q-value overestimation. The dueling architecture in DDDQN enables more effective state-value learning. Additionally, the prioritized experience replay makes it concentrate on critical experiences to accelerate the learning process.

\begin{table}[htbp]
\centering
\caption{Summary of value-based DRL frameworks}
\label{value_based}
\begin{tabular}{p{3.0cm}p{10.5cm}p{1.5cm}}
\hline
DRL Framework & References & Quantity \\
\hline
Q-learning & \cite{zhu2023multi,abdoos2020cooperative,el2013multiagent,chu2016large} & 4\\
DQN & \cite{wang2021adaptive,wang2024large,bie2024multi,liu2023multiple,xu2024graph,wang2021gan,tan2020multi,wu2020multi2,zhang2022neighborhood,lee2019reinforcement,liu2023traffic,wang2020stmarl,fereidooni2025multi,lai2025llmlight,tao2023network} & 15\\
QMIX & \cite{liu2023gplight,bokade2023multi,liu2025globallight} & 3\\
DDQN & \cite{hu2024multi,wang2020large} & 2\\
DDDQN & \cite{jiang2021distributed} & 1\\
\hline
\end{tabular}
\end{table}

\subsubsection{Policy-based DRL methods}
Policy-based DRL methods focus on directly learning the policy that determines the agent’s actions in a given state. Unlike value-based approaches, which learn a value function and derive a policy from it, policy-based methods optimize the policy itself without the need for an explicit value function \citep{li2017deep}. For multi-agent TSC, policy-based DRL methods primarily include Proximal Policy Optimization (PPO) and Counterfactual Multi-Agent (COMA) Policy Gradient. Table \ref{policy_based} provides a summary of the policy-based DRL methods examined in the reviewed studies.

\cite{ren2024two} put forward a two-tier coordinated DRL framework designed for multi-agent TSC. This framework comprises a local cooperation layer and a global cooperation layer. At the local layer, each intersection functions as an independent agent, making decisions based on its immediate traffic conditions. Meanwhile, the global layer ensures the coordination of actions across the entire intersection network. PPO is utilized as the primary reinforcement learning algorithm for agents in both layers.
Similarly, \cite{zhang2022distributed} employed the PPO algorithm, incorporating parameter sharing among multiple agents, to address multi-agent TSC along an arterial corridor. This method accelerates the learning process and minimizes computational overhead.
\cite{luo2024reinforcement} presented a hybrid action space method for optimizing traffic signal control, integrating both discrete and continuous actions. The discrete component manages the selection of traffic light stages, while the continuous component fine-tunes the duration of green light intervals. To efficiently handle these hybrid action spaces, a DRL framework based on PPO was developed. \cite{nie2025cmrm} proposed CMRM, a collaborative MARL method for multi-objective TSC: it augments independent PPO with a Cooperation Enhancement Module (CEM) based on graph attention to share informative neighbor context. \cite{satheesh2025constrained} proposed a constrained DRL method based on PPO. A multi-agent DRL based on PPO actor–critic augmented with a learned Lagrange Cost Estimator that stabilizes updates of the Lagrange multiplier to boost throughput/reduces delay with fewer constraint violations.

\cite{song2024cooperative} suggested a novel DRL framework based on the COMA Policy Gradient method. COMA, a specialized actor-critic approach for multi-agent systems, employs a centralized critic to compute the advantage function for each agent’s actions relative to a counterfactual baseline. The actors function in a decentralized manner, utilizing both local and globally shared information. Furthermore, the framework integrates an innovative scheduler component to improve information sharing among the decentralized actors.

\begin{table}[htbp]
\centering
\caption{Summary of policy-based DRL frameworks}
\label{policy_based}
\begin{tabular}{p{4.0cm}p{10.0cm}p{1.5cm}}
\hline
DRL Framework & References & Quantity \\
\hline
PPO & \cite{ren2024two,zhang2022distributed,luo2024reinforcement,nie2025cmrm,fereidooni2025multi,satheesh2025constrained} & 6\\
COMA Policy Gradient & \cite{song2024cooperative} & 1\\
\hline
\end{tabular}
\end{table}

\subsubsection{Hybrid DRL methods}
Hybrid DRL combines elements from both value-based and policy-based methods, leveraging the strengths of each to mitigate their individual limitations \citep{farazi2021deep}. These hybrid DRL approaches are primarily built on an actor-critic structure. Common frameworks include Advantage Actor-Critic (A2C), Actor-Critic, Deep Deterministic Policy Gradient (DDPG) \citep{wang2024fuel}, Asynchronous Advantage Actor-Critic (A3C), and Soft Actor-Critic (SAC). Table \ref{hybrid_based} provides a summary of hybrid DRL methods applied to the multi-agent TSC research problem.

\cite{yang2023hierarchical} proposed a novel approach that combines multi-granularity information fusion with a mutual information optimization framework for urban traffic signal control. Multi-granularity fusion integrates both current and historical traffic signal states across multiple time steps to capture a comprehensive view of traffic dynamics. GNNs are used to encode state information, leveraging spatial relationships among traffic nodes. The approach employs decentralized execution with centralized training using an actor-critic DRL framework.
\cite{yang2021ihg} devised a multi-agent DRL approach that combines inductive learning with a heterogeneous GNN. In this model, the traffic network is represented as a heterogeneous graph where nodes correspond to different elements, such as vehicles, traffic signals, and intersections. The inductive learning framework can generate embeddings for nodes not encountered during training, which is especially useful in dynamic traffic networks with continuously changing vehicles and conditions. An actor-critic framework supports agent learning in this approach.

\cite{su2023emvlight} employed a multi-agent A2C-based DRL framework for managing dynamic routing and traffic signal control of emergency vehicles (EMVs). This framework integrates dynamic EMV routing with traffic signal preemption, facilitating efficient EMV movement while minimizing the impact on regular traffic. The proposed DRL framework incorporates a policy-sharing mechanism and a spatial discount factor, allowing each traffic signal, functioning as an agent, to make informed decisions based on both local and network-wide information.
\cite{wu2022distributed} introduced two DRL algorithms enhanced by game theory principles: Nash Advantage Actor–Critic (Nash-A2C) and Nash Asynchronous Advantage Actor–Critic (Nash-A3C). Nash-A2C combines the Nash Equilibrium framework with conventional Actor-Critic methods to alleviate congestion and reduce network delays. Nash-A3C builds upon Nash-A2C by introducing asynchronous operations across multiple agents within the traffic network, utilizing a centralized training process paired with decentralized execution.
\cite{chu2019multi} highlighted a multi-agent DRL approach for TSC utilizing a decentralized A2C framework. The approach incorporates neighborhood fingerprints, enabling each agent to consider the recent policies of neighboring agents. This mechanism reduces uncertainty and enhances decision-making despite incomplete information. Additionally, a spatial discount factor is applied to reduce the impact of distant agents, allowing each agent to prioritize local traffic conditions effectively.

\begin{table}[htbp]
\centering
\caption{Summary of hybrid-based DRL frameworks}
\label{hybrid_based}
\begin{tabular}{p{3.0cm}p{10.5cm}p{1.5cm}}
\hline
DRL Framework & References & Quantity \\
\hline
A2C & \cite{su2023emvlight,wu2022distributed,chu2019multi,ma2020feudal} & 4\\
A3C & \cite{wu2022distributed} & 1\\
Actor-Critic & \cite{yang2023hierarchical,yang2021ihg} & 2\\
DDPG & \cite{li2021network,wu2020multi1,jia2025multi,shen2025hierarchical,yoon2025decentralized} & 5\\
SAC & \cite{ge2021multi,zhang2025towards,wang2025towards} & 3\\
\hline
\end{tabular}
\end{table}

\cite{li2021network} outlined a knowledge-sharing DDPG (KS-DDPG) approach for multi-agent TSC. Built on the DDPG algorithm, KS-DDPG employs centralized training with decentralized execution. During training, agents learn from the aggregated experiences of all agents in a centralized manner. However, execution is decentralized, allowing each agent to make independent decisions based on private observations and shared knowledge. The knowledge-sharing component accelerates model convergence without significantly increasing computational demands.
\cite{wu2020multi1} explored a multi-agent recurrent DDPG framework combining DDPG with Long Short-Term Memory (LSTM) to improve the handling of temporal dependencies and partial observability in traffic environments. Both the actor and critic networks in DDPG are equipped with LSTM layers. Simulations validate the model's effectiveness, demonstrating its ability to reduce vehicle and pedestrian congestion more effectively than other methods, particularly in complex urban settings with multiple intersections. \cite{jia2025multi} proposed a hybrid MADRL traffic-signal framework that fuses spatio-temporal attention (GAT for spatial deps + LSTM for temporal trends) with hierarchical control (sub-region agents + a global coordinator) under CTDE using multi-agent DDPG learning framework. The results report 25\% lower average waiting time and the highest throughput versus fixed-time, actuated, Max-Pressure, DQN, PPO, and vanilla MADDPG baselines. 

\cite{ge2021multi} designed a multi-agent DRL approach based on SAC, tailored for multi-agent environments with entropy maximization to maintain a balance between exploration and exploitation. The approach incorporates a multi-view encoder to process state inputs from diverse perspectives: a 1D vector represents local intersection data, while a 2D state representation, processed through a CNN, captures spatial traffic dynamics. Furthermore, a transfer learning mechanism with cooperative guidance enables agents to develop generalizable skills while adapting to the unique characteristics of specific intersections. \cite{zhang2025towards} proposed an M$^{2}$SAC, a cooperative masked SAC for corridor-level signal control, where a hybrid actor network generates a Gaussian-sampled phase mask that can defer the last phase to the next cycle and then picks green-time splits via a masked softmax function. Ablation studies show the introduced mask mechanism yields extra gains.

\subsection{Summary}
The majority of existing multi-agent TSC research relies on SUMO or CityFlow for traffic environment simulation. Most prior studies design reward functions to reduce vehicle-centric congestion or delay, while only limited attention has been given to human-centric objectives that reflect individual traveler experience across transport modes. In terms of representation learning, existing studies have predominantly relied on GNNs and CNNs, whereas hypergraph-based formulations have received comparatively little attention. In addition, current approaches generally fall into three categories: value-based, policy-based, and hybrid DRL methods, with DQN, PPO, and actor-critic variants being the most commonly adopted frameworks. Overall, recent research has largely focused on adapting and refining established DRL methods for traffic signal optimization rather than addressing the combined challenges of multimodal human-centric control and unified higher-order spatio-temporal modeling.

Despite this progress, three important gaps remain. First, most existing studies optimize vehicle-centric indicators such as queue length, delay, or waiting time, which do not explicitly reflect passenger-level impacts across different transport modes. Second, although some studies incorporate graph-based spatial modeling, most rely on pairwise adjacency structures that are limited in capturing higher-order dependencies among multiple intersections and across time. Third, existing spatio-temporal approaches often treat spatial and temporal relationships as separate modeling components rather than learning them within a unified representation. These limitations are particularly important for multimodal corridor control, where high-occupancy public transport and cross-intersection temporal interactions play a central role. To address these gaps, this paper proposes a human-centric multi-agent DRL framework based on a unified spatio-temporal dual-stage hypergraph representation for multimodal corridor traffic signal control.

\section{Methodology}
\label{method}
This section first provides an overview of the proposed methodology. It then presents a detailed description of each key component, including the spatio-temporal hypergraph, dual-stage hypergraph attention mechanism, state representation, action space, reward design, and the MARL training process.

\subsection{Overview}
\begin{figure}[htbp]
\begin{center}
\includegraphics[width=0.95\textwidth]{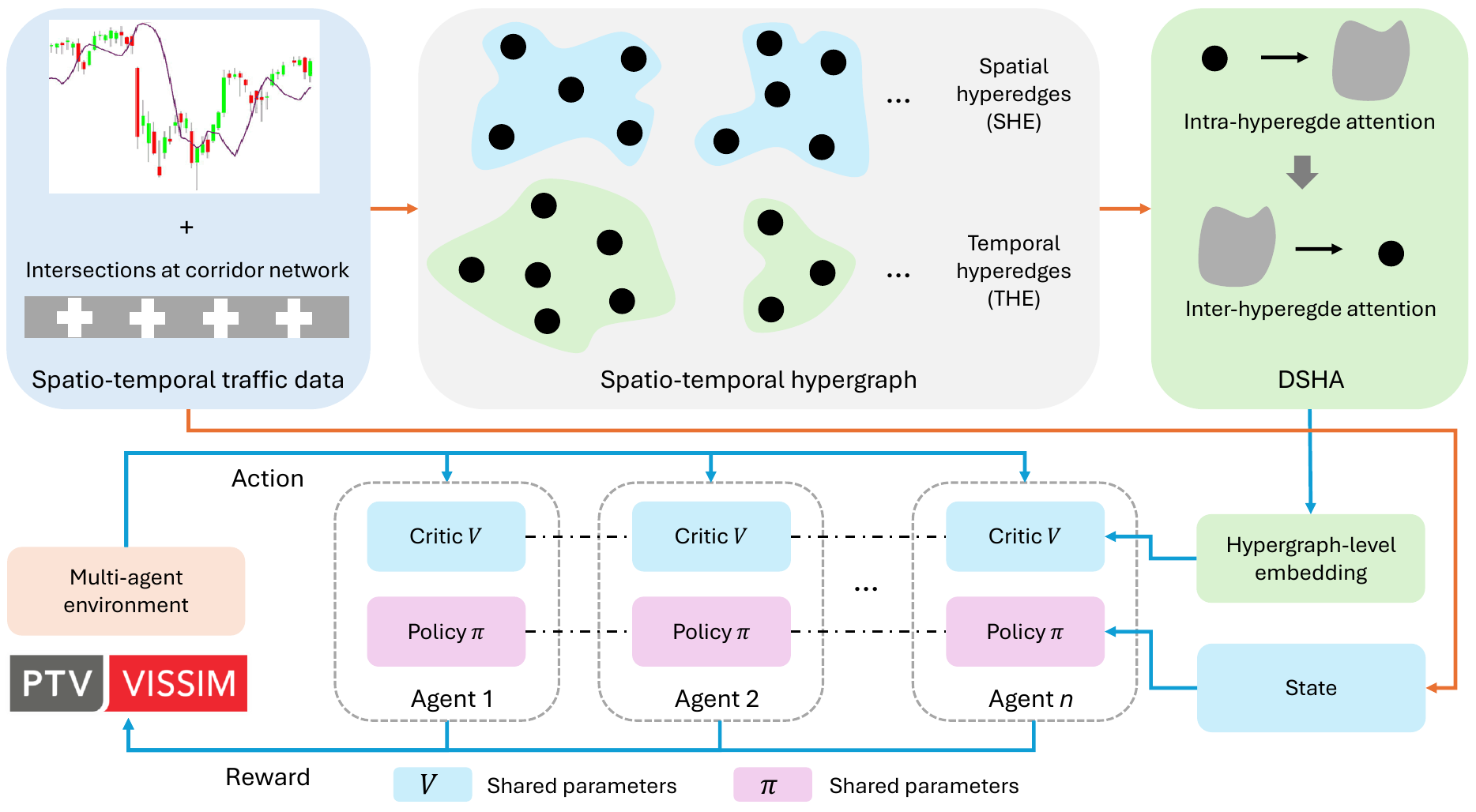}
\caption{Overview of STDSH-MARL: DSHA learns a hypergraph-level embedding from spatial and temporal hyperedges for a centralized critic, while decentralized agents act from local state representations in a VISSIM-simulated corridor.}
\label{omethod}
\end{center}
\end{figure}

Figure~\ref{omethod} provides an overview of the proposed methodology. The inputs comprise spatio-temporal traffic data over a specified time window and the corridor network topology. The framework has two parallel streams: (i) construction of a spatio-temporal hypergraph followed by the dual-stage hypergraph attention (DSHA) module to learn a hypergraph-level embedding, and (ii) formation of the agent state representation for control. The spatio-temporal hypergraph consists of nodes and hyperedges, where the hyperedges include spatial and temporal hyperedges. DSHA is the core novel component and integrates two attention mechanisms: intra-hyperedge attention and inter-hyperedge attention. Intra-hyperedge attention models interactions among nodes within the same hyperedge, whereas inter-hyperedge attention learns the relative importance of different hyperedges. The resulting hypergraph-level embedding is fed into the critic network, while the state representation is fed into the policy network. The multi-agent, multimodal traffic environment is simulated in PTV VISSIM.

\subsection{Spatio-temporal hypergraph}
A hypergraph is a generalization of a conventional graph that allows each edge, called a \emph{hyperedge}, to connect more than two nodes.
Formally, a spatio-temporal hypergraph is defined as:
\begin{equation}
\mathcal{G} = (\mathcal{V}, \mathcal{E}),
\end{equation}
and
\begin{equation}
\mathcal{E} = \mathcal{E}_{s} \cup  \mathcal{E}_{t},
\end{equation}
where 
\(\mathcal{V} = \{v_1, v_2, \ldots, v_N\}\) denotes the set of \textit{nodes} and \(\mathcal{E}\) denotes the set of hyperedges, it includes spatial hyperedges \(\mathcal{E}_{s}\) and temporal hyperedges \(\mathcal{E}_{t}\). where each hyperedge connects one or more nodes.
\(\mathcal{E}_{s} = \{e_{1}^{(s)}, e_{2}^{(s)}, \ldots, e_{M}^{(s)}\}\) and \(\mathcal{E}_{t} = \{e_{1}^{(t)}, e_{2}^{(t)}, \ldots, e_{P}^{(t)}\}\).

The relationships between nodes and hyperedges are encoded in the \textit{node–hyperedge incidence matrix} 
\(\mathbf{H} \in \mathbb{R}^{N \times (M+P)}\), as defined by
\begin{equation}
H_{ij} =
\begin{cases}
1, & \text{if } v_i \in e_j, e_j \in \mathcal{E},\\
0, & \text{otherwise.}
\end{cases}
\end{equation}

By extending the hypergraph formulation to corridor-level traffic signal control, we consider a corridor network with $n$ intersections. Let $t$ denote the number of historical time steps and $T$ the current time step. The temporal index set over the rolling time window is defined as \((T-t+1,\cdots ,T-1,T)\). The node set, which represents the intersections in the corridor, is given by
\begin{equation}
\mathcal{V}
=
\left\{
v_{i,\tau}
\mid
i=1,\ldots,n,\;
\tau=1,\ldots,t
\right\},
\end{equation}
where \(v_{i,\tau}\) represents intersection \(i\) at historical time step
\(\tau\). Consequently, the total number of spatio-temporal nodes is
\begin{equation}
N = |\mathcal{V}| = nt.
\end{equation}

For the spatial hyperedges, we set $M=t$, meaning that $t$ spatial hyperedges are constructed. Specifically, at each timestamp within the rolling time window, all intersections are included in one spatial hyperedge to capture spatial interdependencies at that time step. The intersections are indexed according to their physical order along the corridor, from upstream to downstream. Therefore, although each spatial hyperedge contains all intersections at a given time step, the node sequence preserves the corridor topology and enables the attention mechanism to learn directional upstream-downstream dependencies.
For the temporal hyperedges, we set $P=n$, meaning that $n$ temporal hyperedges are constructed. Specifically, for each intersection, all timestamps within the rolling time window are included in one temporal hyperedge to capture temporal interdependencies of that intersection across time. All intersections or time steps are included in each corresponding hyperedge so that the dual-stage hypergraph attention mechanism can automatically learn the most relevant dependencies, without requiring manual specification of which elements should be grouped together. This avoids the need to exhaustively construct a large number of fine-grained hyperedges, which would require substantial prior knowledge and increase modeling complexity. In this way, the proposed design enables temporal and spatial dependencies to be represented within a unified spatio-temporal hypergraph, rather than being handled as separate modules.

\subsection{Dual-stage hypergraph attention}
Let $\mathbf{X}^{\mathrm{raw}}\in\mathbb{R}^{N\times d_0}$
denote the raw node feature matrix, where $d_0$ is the dimension of the raw traffic features. Let
$\mathbf{H}\in\{0,1\}^{N\times E}$
denote the node--hyperedge incidence matrix, where $E=M+P$.

\paragraph{Spatio-temporal positional encoding.}
Before applying hypergraph attention, each node feature is augmented with spatial and temporal positional information. Each spatio-temporal node $q$ corresponds to intersection $i\in\{1,\ldots,n\}$ and time step $\tau\in\{1,\ldots,t\}$. Their normalized positional values are defined as
\begin{equation}
p_i^{s}=\frac{i-1}{n-1},
\qquad
p_{\tau}^{t}=\frac{\tau-1}{t-1}.
\end{equation}

The position-augmented feature of node $q$, associated with the intersection--time pair $(i,\tau)$, is given by
\begin{equation}
\bar{\mathbf{x}}_{q}
=
\mathbf{x}^{\mathrm{raw}}_{q}
\mathbin{\|} p_i^{s}
\mathbin{\|} p_{\tau}^{t},
\end{equation}
where $\|$ denotes feature concatenation. The augmented node feature matrix $\bar{\mathbf{X}}$ is subsequently mapped to the model dimension $d$ using a multilayer perceptron:
\begin{equation}
\mathbf{X}
=
\operatorname{MLP}\!\left(\bar{\mathbf{X}}\right)
\in\mathbb{R}^{N\times d}.
\end{equation}

We use $K$ attention heads with per-head dimension $d_h=d/K$. For head $h\in\{1,\ldots,K\}$, the head-specific node embeddings are obtained as
\begin{equation}
\mathbf{X}^{(h)}
=
\mathbf{X}\mathbf{W}^{(h)}
\in\mathbb{R}^{N\times d_h},
\qquad
\mathbf{W}^{(h)}
\in\mathbb{R}^{d\times d_h},
\end{equation}
where $\mathbf{W}^{(h)}$ is a learnable projection matrix.

We employ a dual-stage attention mechanism consisting of: 
(A) intra-hyperedge attention and 
(B) inter-hyperedge attention. 
Intra-hyperedge attention assigns importance scores to the member nodes of each hyperedge and aggregates them to form a hyperedge embedding. Inter-hyperedge attention then evaluates the hyperedges incident to each node and determines the relative importance of spatial and temporal dependencies.

\paragraph{Stage A: Intra-hyperedge attention.}
For each attention head, the embedding of node $i$ is defined as
\begin{equation}
\mathbf{x}_i^{(h)}
=
\left(\mathbf{X}_{i:}^{(h)}\right)^{\top}
\in\mathbb{R}^{d_h}.
\end{equation}

A scalar importance score is then computed for each node:
\begin{equation}
s_i^{(h)}
=
\mathbf{a}^{(h)\top}\mathbf{x}_i^{(h)},
\qquad
\mathbf{a}^{(h)}
\in\mathbb{R}^{d_h},
\end{equation}
where $\mathbf{a}^{(h)}$ is a learnable attention vector.

The attention weight of node $i$ within hyperedge $e$ is calculated using a masked softmax over the member nodes of that hyperedge:
\begin{equation}
\alpha_{ie}^{(h)}
=
\frac{
\exp\!\left(s_i^{(h)}-m_e^{(h)}\right)H_{ie}
}{
\displaystyle
\sum_{j=1}^{N}
\exp\!\left(s_j^{(h)}-m_e^{(h)}\right)H_{je}
},
\qquad
m_e^{(h)}
=
\max_{j:\,H_{je}=1}s_j^{(h)}.
\label{eq:alpha}
\end{equation}

The embedding of hyperedge $e$ is obtained by aggregating its member-node embeddings:
\begin{equation}
\mathbf{z}_e^{(h)}
=
\sum_{i\in\mathcal{N}(e)}
\alpha_{ie}^{(h)}
\mathbf{x}_i^{(h)}
\in\mathbb{R}^{d_h}.
\label{eq:edge-agg}
\end{equation}

\paragraph{Stage B: Inter-hyperedge attention}
Each head scores hyperedge embeddings and normalizes across the incident hyperedges of each node, as shown by
\begin{equation}
t_e^{(h)} = \mathbf{b}^{(h)\top}\mathbf{z}_{e}^{(h)},\qquad \mathbf{b}^{(h)}\in\mathbb{R}^{d_h},
\end{equation}
and
\begin{equation}
\beta_{ie}^{(h)} =
\frac{\exp\!\big(t_e^{(h)}-m_i^{(h)}\big)\, H_{ie}}
{\displaystyle \sum_{e'=1}^{E} \exp\!\big(t_{e'}^{(h)}-m_i^{(h)}\big)\, H_{ie'}},
\qquad
m_i^{(h)}=\max_{e':\,H_{ie'}=1} t_{e'}^{(h)}.
\label{eq:beta-H}
\end{equation}
In Eq.~\eqref{eq:beta-H} we write the denominator as \(\sum_{e'=1}^{E} \exp(t_{e'}^{(h)}-m_i^{(h)})\,H_{ie'}\). Here \(e'\) is a \emph{dummy summation index} (also called a bound variable) that ranges over all hyperedges; it is used to distinguish the variable of summation from the fixed hyperedge \(e\) appearing in the numerator.
The updated node representation for head $h$ is obtained by aggregating hyperedge embeddings back to nodes, as shown by
\begin{equation}
\tilde{\mathbf{x}}_{i}^{(h)} \;=\; \sum_{e=1}^{E} \beta_{ie}^{(h)}\, H_{ie}\, \mathbf{z}_e^{(h)}
\;\in\; \mathbb{R}^{d_h}, 
\qquad i=1,\dots,N,\; h=1,\dots,K,
\label{eq:node-agg}
\end{equation}
and
\begin{equation}
\tilde{\mathbf{X}}^{(h)} \;=\;
\begin{bmatrix}
(\tilde{\mathbf{x}}_{1}^{(h)})^{\!\top}\\
\vdots\\
(\tilde{\mathbf{x}}_{N}^{(h)})^{\!\top}
\end{bmatrix}
\in \mathbb{R}^{N\times d_h},
\qquad h=1,\dots,K.
\label{eq:node-stack}
\end{equation}
We then concatenate per-head outputs, as shown by
\begin{equation}
\tilde{\mathbf{X}} = \big\|_{h=1}^{H}\tilde{\mathbf{X}}^{(h)}
\;\in\; \mathbb{R}^{N\times d_{\text{cat}}}, 
\qquad d_{\text{cat}}=\sum_{h=1}^{K} d_h ,
\label{eq:node-concat}
\end{equation}
and project to the model width, as formulated by
\begin{equation}
\hat{\mathbf{Y}} = \tilde{\mathbf{X}}\mathbf{W}_o + \mathbf{b}_o
\in \mathbb{R}^{N\times d_{\text{model}}}.
\label{eq:node-proj}
\end{equation}
A permutation-invariant readout produces a hypergraph-level embedding $\mathbf{g}\in\mathbb{R}^{d_{\mathrm{model}}}$.
Consistent with our implementation, we use max pooling:
\begin{equation}
\mathbf{g} = \operatorname*{MaxPool}_{i=1,\dots,N}\,\hat{\mathbf{Y}}_{i:}
\;\in\; \mathbb{R}^{d_{model}}.
\label{eq:readout-max}
\end{equation}
Unlike conventional graph attention mechanisms that capture pairwise interactions, the proposed dual-stage hypergraph attention explicitly models higher-order spatial and temporal dependencies, enabling more expressive representation learning for corridor-level traffic coordination.

\subsection{State representation}
\begin{figure}[htbp]
\begin{center}
\includegraphics[width=0.7\textwidth]{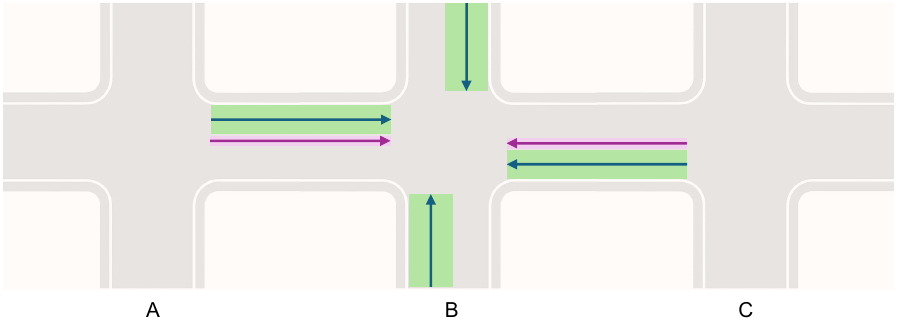}
\caption{Coverage of the approaching lanes for the example intersection B among a corridor network. Green stands for the vehicle lanes and pink denotes tram lanes. The schematic assumes left-hand traffic.}
\label{approachinglanes}
\end{center}
\end{figure}

For each intersection, we adopt a lane-level state representation. Specifically, we extract a compact set of high-level traffic descriptors from each approaching lane (Figure~\ref{approachinglanes}), which are sufficiently representative while avoiding excessive computational and memory overhead. The extracted information covers four aspects: vehicle count, passenger count, queue length, and speed. The detailed lane-wise features are summarized in Table \ref{feature}. Overall, we use four feature categories to characterize the prevailing traffic conditions: (i) approaching vehicle demand, (ii) passenger profiles, (iii) queue length, and (iv) speed profiles. For each category, we compute both aggregate features and mode-specific features (e.g., for buses, trams, and private vehicles). Finally, we concatenate the features from all approaching lanes into a single vector, which serves as the state representation of the intersection (i.e., the node state).

\begin{table}[htbp]
\centering
\caption{Features to define the state representation}
\label{feature}
\begin{threeparttable}
\begin{tabular}{p{3.5cm}p{10cm}}
\hline
Type & Features \\
\hline
Vehicle number & Total vehicle number, number of buses, number of trams, number of private vehicles.\\
\hline
Passenger number & Total passenger number, passenger number for buses, passenger number for trams, passenger number for private vehicles.\\
\hline
Queue length & Total queue length, queue length for buses, queue length for trams, queue length for private vehicles.\\
\hline
Speed & Average speed, average speed for buses, average speed for trams, average speed for private vehicles.\\ 
\hline
Current signal phase & One-hot embedding of current signal phase for corresponding intersection.\\
\hline
\end{tabular}
        \textit{Notes:} Vehicle in this study also includes trams. Passengers in private vehicles also account for the driver.
\end{threeparttable}
\end{table}

\subsection{Action space}
\begin{figure}[htbp]
\begin{center}
\includegraphics[width=0.98\textwidth]{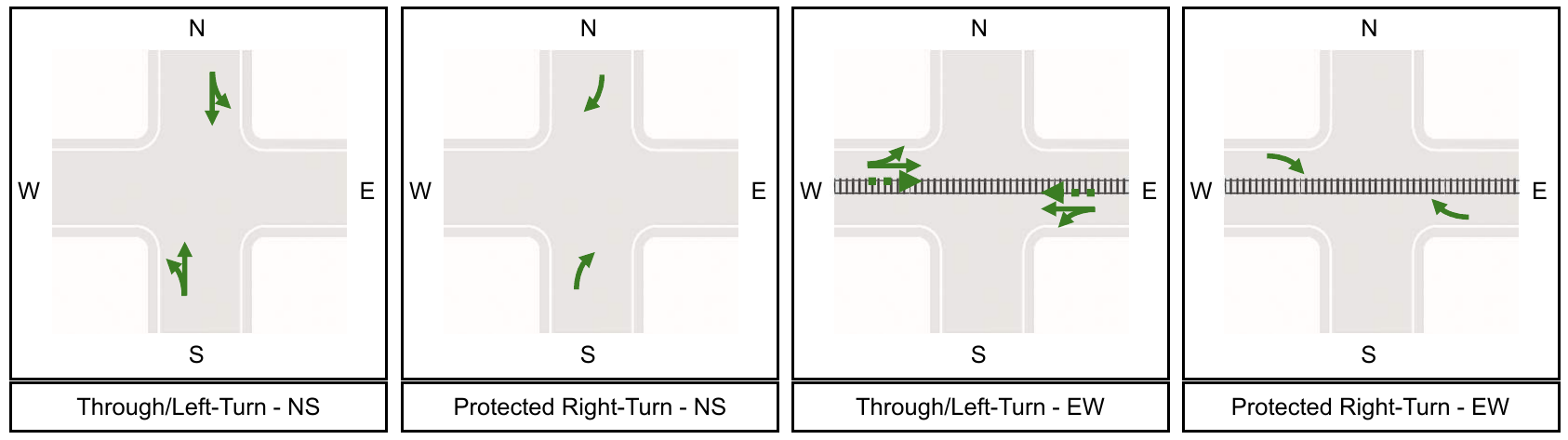}
\caption{Phase configurations for action selection.}
\label{phaes}
\end{center}
\end{figure}

Compared with most studies that use either the next phase or the green duration as the action, we jointly select the next phase and its corresponding green time. Phase selection is drawn from a set of four configurations, as shown in Figure~\ref{phaes}. For each direction, two movement patterns are available: (i) through- and left-turn movements, and (ii) protected right-turn movements, resulting in four phase configurations in total. At each decision step, the agent chooses one phase configuration for the next phase, and consecutive selection of the same configuration is prohibited to avoid repeated phases.

The green time duration is an integer value selected from 8 seconds to 45 seconds, there are 38 possible green time actions. Between each phase transition, we have designed 3-second amber time and 2-second all-red-clearance time to make safer phase transitions. We convert the whole action by combining the phase configuration and the green duration as a discrete space with a dimension of 152, i.e., $38 \times 4 = 152$.

\subsection{Reward mechanism}
The human-centric objective in this study is defined at the passenger level rather than the vehicle level. Each traveler contributes equally to the reward, regardless of whether they are in a private vehicle, bus, or tram. As a result, high-occupancy public transport may receive implicit operational preference because delaying one bus or tram affects more travelers, not because the reward assigns explicit mode-specific priority weights.

Let $\mathcal{U}_{i,t}$ denote the set of vehicles within the control region of intersection $i$ at time step $t$, and let $q_{u,t}$ denote the number of travelers carried by vehicle $u \in \mathcal{U}_{i,t}$. For private vehicles, $q_{u,t}$ includes both the driver and passengers, while for buses and trams, $q_{u,t}$ denotes the onboard passenger count. A vehicle is considered delayed when its instantaneous speed falls below a predefined threshold, as shown by
\begin{equation}
d_{u,t} =
\begin{cases}
1, & \text{if } v_{u,t} < \theta_v,\\
0, & \text{otherwise},
\end{cases}
\label{eq:delay_indicator}
\end{equation}
where $v_{u,t}$ denotes the instantaneous speed of vehicle $u$ at time step $t$, and $\theta_v$ is the delay threshold, set to $0.02$ km/h in this study. The resulting delayed-passenger count is used as a proxy for passenger-level traffic impact. It represents the number of passengers in effectively stationary vehicles at a given time, rather than a direct measure of accumulated passenger delay or total travel time.

The local delayed-traveler count at intersection $i$ and time step $t$ is defined as
\begin{equation}
N_{i,t} = \sum_{u \in \mathcal{U}_{i,t}} q_{u,t} \, d_{u,t}.
\label{eq:local_delayed_passengers}
\end{equation}

The corresponding network-wide delayed-traveler count is
\begin{equation}
\hat{N}_t = \sum_{i=1}^{n} N_{i,t},
\label{eq:global_delayed_passengers}
\end{equation}
where $n$ is the total number of intersections in the corridor network.

Accordingly, the reward for agent $i$ at decision step $k$ is defined as
\begin{equation}
r_{i,k} = - \left(
\frac{w_1}{m} \sum_{t=1}^{m} N_{i,t}
+
\frac{w_2}{m} \sum_{t=1}^{m} \hat{N}_t
\right),
\label{eq:reward}
\end{equation}
where $m$ is the control duration in seconds, and $w_1$ and $w_2$ are weighting parameters that balance local and network-wide delayed-traveler impacts, respectively. Note that no explicit mode-specific weighting is introduced in this study.

\subsection{MARL learning}
We formulate corridor-level traffic signal control as a Markov Decision Process (MDP) \citep{zhang2026realistic} under a cooperative multi-agent reinforcement learning (MARL) framework with agents $\mathcal{I}=\{1,\dots,n\}$, where each agent represents an intersection. At time $t$, agent $i$ observes a local state $o_t^{(i)}\in\mathbb{R}^{d_o}$ and selects a discrete action $a_t^{(i)}\in\mathcal{A}$ (e.g., a phase-duration option). Let $\mathbf{o}_t=(o_t^{(1)},\dots,o_t^{(n)})$ and
$\mathbf{a}_t=(a_t^{(1)},\dots,a_t^{(n)})$ denote the joint observation and joint action, respectively. The environment provides a reward $r_t$, transitions to the next joint observation $\mathbf{o}_{t+1}=(o_{t+1}^{(1)},\dots,o_{t+1}^{(n)})$, and the objective is to maximize the expected discounted return with discount factor $\gamma \in \left ( 0,1 \right )$. We adopt a CTDE paradigm instantiated with an on-policy, actor–critic PPO framework, where each agent’s actor produces its action from local observations while a critic is trained to evaluate the joint behavior during training.

\paragraph{Decentralized execution with parameter sharing}
We adopt a parameter-sharing scheme in which all agents use a shared actor $\pi_\theta$. Under this setting, the joint policy factorizes as
\begin{equation}
\pi_\theta(\mathbf{a}_t\mid\mathbf{o}_t)
=\prod_{i=1}^{n}\pi_\theta\!\big(a_t^{(i)}\mid o_t^{(i)}\big),
\label{eq:factored-policy}
\end{equation}
enabling fully decentralized action selection at execution time. In implementation, $\pi_\theta$ is a two-layer MLP that maps each agent’s flattened local observation to a vector of logits over the discrete action set $\mathcal{A}$.

\paragraph{Centralized training via a hypergraph critic}
To facilitate credit assignment, we employ a centralized value function
$V_\psi(\mathbf{g}_t)$, where $\mathbf{g}_t\in\mathbb{R}^{d}$ is a
\emph{hypergraph-level} embedding produced by the dual-stage hypergraph
attention module (Eq.~\eqref{eq:readout-max}). The critic $V_\psi$ is implemented as a two-layer MLP that maps $\mathbf{g}_t$ to a scalar value estimate.

\paragraph{On-policy updates with PPO}
From a rollout $\{(\mathbf{o}_t,\mathbf{a}_t,r_t)\}_{t=0}^{T-1}$ we compute
the discounted returns and advantages as
\begin{equation}
\hat{R}_t=\sum_{k=0}^{T-1-t}\gamma^{k} r_{t+k},
\label{eq:marl-rew}
\end{equation}
and
\begin{equation}
\hat{A}_t=\hat{R}_t - V_\psi(\mathbf{g}_t^{\varphi}),
\label{eq:marl-adv}
\end{equation}
where $V_\psi(\mathbf{g}_t^{\varphi})$ is the centralized value estimate based on the hypergraph embedding at time $t$. We then optimize the actor using the clipped PPO objective (evaluated per time step and aggregated over the batch):
\begin{equation}
\mathcal{L}_{\mathrm{actor}}(\theta)
=-\,\mathbb{E}\!\left[
\min\!\left(
\rho_t(\theta)\,\hat{A}_t,\;
\mathrm{clip}\!\big(\rho_t(\theta),1{-}\epsilon,1{+}\epsilon\big)\,\hat{A}_t
\right)\right]
-\alpha_{\mathrm{entropy}}\;\mathbb{E}\big[\mathcal{H}(\pi_\theta(\cdot\mid o_t^{(i)}))\big],
\label{eq:marl-ppo}
\end{equation}
and
\begin{equation}
\rho_t(\theta)=\exp\!\big(\log\pi_\theta-\log\pi_{\theta_{\text{old}}}\big),
\label{eq:marl-ppo-2}
\end{equation}
where $\alpha_{\mathrm{entropy}}>0$ is the entropy-regularization coefficient that encourages exploration by rewarding higher policy entropy; larger values yield more stochastic policies (better exploration, slower convergence), while smaller values produce greedier policies (faster convergence, higher risk of premature collapse). The critic loss is formulated as
\begin{equation}
\mathcal{L}_{\mathrm{critic}}(\psi)=\tfrac{1}{2}\,\mathbb{E}\big[(\hat{R}_t - V_\psi(\mathbf{g}_t^{\varphi}))^2\big],
\label{eq:marl-vloss}
\end{equation}
and
\begin{equation}
\mathcal{L}_{\mathrm{hypergraph}}(\varphi)
= \mathcal{L}_{\mathrm{critic}}.
\label{eq:ent}
\end{equation}
For each update, we iterate over shuffled mini-batches for multiple PPO epochs and apply optional global-norm clipping to stabilize gradients. In implementation, we perform gradient-based updates by minimizing the actor loss $\mathcal{L}_{\mathrm{actor}}(\theta)$, which is the negative of the clipped PPO surrogate augmented with an entropy bonus (i.e., equivalently maximizing the surrogate objective). The critic (and hypergraph encoder) is trained by minimizing the value regression loss $\mathcal{L}_{\mathrm{critic}}(\psi)$. The hypergraph attention module is trained jointly with the critic loss by
optimizing the hypergraph parameters $\varphi$.

At each time step, only agents whose local trigger is active resample a new action from $\pi_\theta$; the remaining agents retain their previous action (phase), thereby reducing unnecessary switching. Execution remains fully decentralized via Eq.~\eqref{eq:factored-policy}, while the centralized critic exploits global hypergraph context during training.

\section{Experiments and results}
\label{experiment}
\subsection{Experimental setup}
\begin{figure}[htbp]
\begin{center}
\includegraphics[width=1\textwidth]{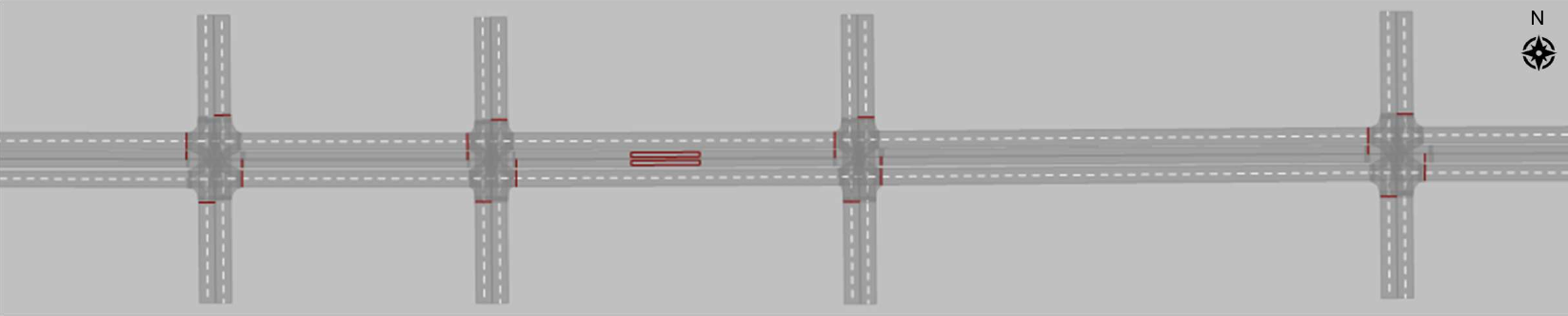}
\caption{Partial view of the constructed corridor network in VISSIM for experimental evaluation.}
\label{network}
\end{center}
\end{figure}

We construct a synthetic road corridor in Australia under a left-hand driving environment consisting of six signalized intersections. Each intersection has four approaches, and each approach comprises two incoming lanes. A tram track runs along the corridor, and three tram stops are modeled within the network. The traffic environment is simulated in PTV VISSIM\footnote[6]{\url{https://www.ptvgroup.com/en/products/ptv-vissim}}
, a microscopic traffic simulation platform. Figure~\ref{network} shows the (partial) corridor network used in the experiments.

Passenger occupancy is randomly generated according to vehicle type using discrete uniform distributions. Private vehicles carry 1–5 occupants, buses carry 15–50 passengers, and trams carry 20–80 passengers. Other vehicle types are assumed to carry only one occupant.

The proposed model is evaluated under five independent scenarios. Scenario~1 represents an off-peak period with low demand. Scenario~2 captures the transition from off-peak to peak conditions with moderate demand. Scenario~3 simulates rush-hour traffic with high demand. Scenarios~4 and~5 represent school-period conditions: Scenario~4 corresponds to the morning school peak with increased inbound traffic toward a specific area, whereas Scenario~5 corresponds to the afternoon school peak with increased outbound traffic away from that area. Detailed traffic-demand settings for all five scenarios are provided in the accompanying online repository\footnote[7]{\url{https://1drv.ms/f/c/aa681a6090739da3/IgBAy4v404htQb3EmE0yF072AQp1VV9XlbP1Lw008mgqDJ8?e=jXhkZQ}}. The 14 traffic zones are indexed in clockwise order, with Zone~1 corresponding to the north approach of the westernmost intersection along the corridor.
All models are implemented in Python using TensorFlow, and experiments are conducted on a Windows workstation equipped with an AMD EPYC 7702 CPU (64 cores, 1996~MHz). To ensure reproducibility and consistent stochastic conditions across experiments, the random seed in PTV VISSIM was fixed at 101 for all simulation runs.

\subsubsection{Hyperparameters}
The main hyperparameter settings of the proposed STDSH-MARL model are summarized in Table~\ref{tab:hyperparameters}. In the dual-stage hypergraph attention module, the number of spatial hyperedges \(M\) and temporal hyperedges \(P\) are set to 5 and 6, respectively, to capture spatial and temporal dependency structures. The number of attention heads \(K\) is set to 4, allowing the model to learn multiple representation subspaces. The model dimension \(d_{\mathrm{model}}\) is set to 64, while the hidden-layer dimension is set to 128, providing sufficient representation capacity while keeping the model computationally efficient.

For reinforcement learning training, the batch size is set to 64. The learning rates for the actor and critic networks are set to 0.0001 and 0.0003, respectively. The discount factor \(\gamma\) is set to 0.99, which allows the agent to consider long-term rewards. The PPO clipping range \(\epsilon\) is set to 0.1 to restrict overly large policy updates and improve training stability. The entropy coefficient \(\alpha_{\mathrm{entropy}}\) is set to 0.001 to encourage exploration during policy learning. The reward weighting parameters $w_1$ and $w_2$ are set to 0.6 and 0.4, respectively. The model is trained for 300 episodes.

\begin{table}[htbp]
\centering
\caption{Hyperparameter settings used in the proposed STDSH-MARL model.}
\label{tab:hyperparameters}
\begin{tabular}{p{6cm}p{1.5cm}p{6cm}p{1.5cm}}
\hline
Hyperparameter & Value & Hyperparameter & Value \\
\hline
$M$ - Number of spatial hyperedges & 5 & Batch size & 64 \\
$P$ - Number of temporal hyperedges & 6 & Actor learning rate & 0.0001 \\
$K$ - Number of heads & 4 & Critic learning rate & 0.0003 \\
$d_{model}$ - Model dimension & 64 & Number of episodes & 300 \\
Hidden dimension & 128 & $\epsilon$ - Clipping range & 0.1 \\
$\gamma$ - Discount factor & 0.99 & $\alpha_{\mathrm{entropy}}$ - Entropy coefficient & 0.001 \\
$w_1$ - Weighting parameter & 0.6 & $w_2$ - Weighting parameter & 0.4\\
\hline
\end{tabular}
\end{table}

\subsubsection{Evaluation metrics}
We evaluate model performance using the following network-wide metrics: (1) Average number of passengers experiencing delay (ANP), which quantifies the average number of delayed passengers per second over the test horizon, aggregated across all transportation modes in the network; (2) Average vehicle queue length (AQL), defined as the average number of road vehicles queued per second upstream of intersections across the network (excluding trams); (3) Average waiting time for buses (AWT (bus)), i.e., the mean time (s) a bus spends waiting to traverse the corridor; and (4) Average waiting time for trams (AWT (tram)), i.e., the mean time (s) a tram spends waiting to traverse the corridor. The two AWT metrics primarily capture the effectiveness of public-transportation prioritization.

\subsubsection{Benchmarking}
To evaluate the performance of the proposed STDSH-MARL, we benchmark it against a range of baseline methods, including fixed-time signal control and several DRL-based ATSC approaches, namely MADQN, MADDQN, MAA2C, MAPPO, and CMRM.
\begin{itemize}
  \item \textbf{Fixed Signal with Webster's Formula (FS-WF)}: FS-WF employs Webster's Formula to estimate the optimal cycle time, followed by green splitting to allocate green durations for each phase. Both the optimal cycle time estimation and green splitting are based on traffic demand.
  \item \textbf{Actuated} \citep{yazdani2023intelligent}: Actuated signal control adjusts signal phases and green durations according to real-time traffic demand detected at each intersection. Vehicle detectors are used to determine whether a phase should be extended, terminated, or skipped, subject to predefined minimum and maximum green time constraints.
  \item \textbf{Max-Pressure (MP) Control} \citep{ramadhan2020application}: MP control selects the phase with the highest pressure, defined by the queue imbalance between upstream and downstream movements. It prioritizes highly congested movements to improve traffic throughput without requiring model training.
  \item \textbf{MADQN} \citep{wei2018intellilight}: This baseline formulates traffic signal control with a discrete action space using a DQN-based approach. It adopts a fully Decentralized Training and Decentralized Execution (DTDE) paradigm, where each agent learns and operates independently in the multi-agent environment.
  \item \textbf{MADDQN} \citep{wang2020large}: This baseline extends MADQN by incorporating a DDQN as the underlying model to mitigate the Q-value overestimation issue inherent in standard DQN. It likewise follows the same DTDE paradigm as MADQN.
  \item \textbf{MAA2C} \citep{su2023emvlight}: MAA2C is fundamentally developed on the well-established A2C framework, which provides stable and efficient learning for policy-based reinforcement learning. It adopts a CTDE paradigm for multi-agent environments.
  \item \textbf{MAPPO}: This baseline applies standard multi-agent PPO without incorporating hypergraph-based representation learning. It is included to isolate and quantify the benefit of introducing hypergraph embeddings, by directly comparing performance with and without the hypergraph module.
  \item \textbf{CMRM} \citep{nie2025cmrm}: It combines a GAT with PPO for multi-agent traffic signal optimization. This baseline excludes hypergraph embeddings, serving as a standard graph-based alternative to assess the added value of hypergraph representations over conventional graph embeddings. It adopts a CTDE paradigm.
\end{itemize}

\begin{figure}[htbp]
\begin{center}
\includegraphics[width=0.75\textwidth]{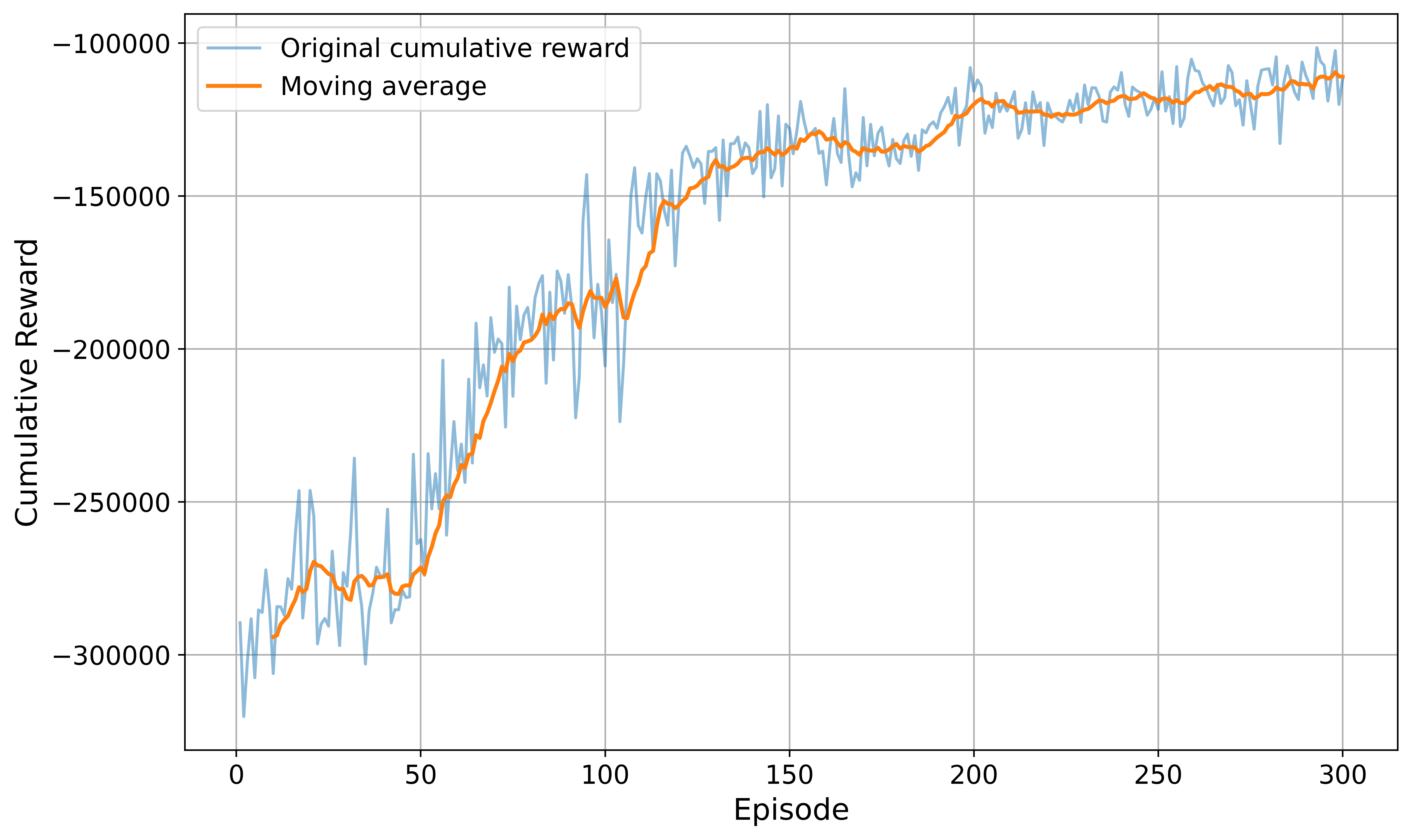}
\caption{Training reward curve over the episodes. The light blue line represents the original cumulative reward, while the dark orange line represents the 10-episode moving average.}
\label{rewardcurve}
\end{center}
\end{figure}

\begin{figure}[htbp]
\centering
\subfloat[Number of delayed passengers.]{%
  \includegraphics[width=0.95\textwidth]{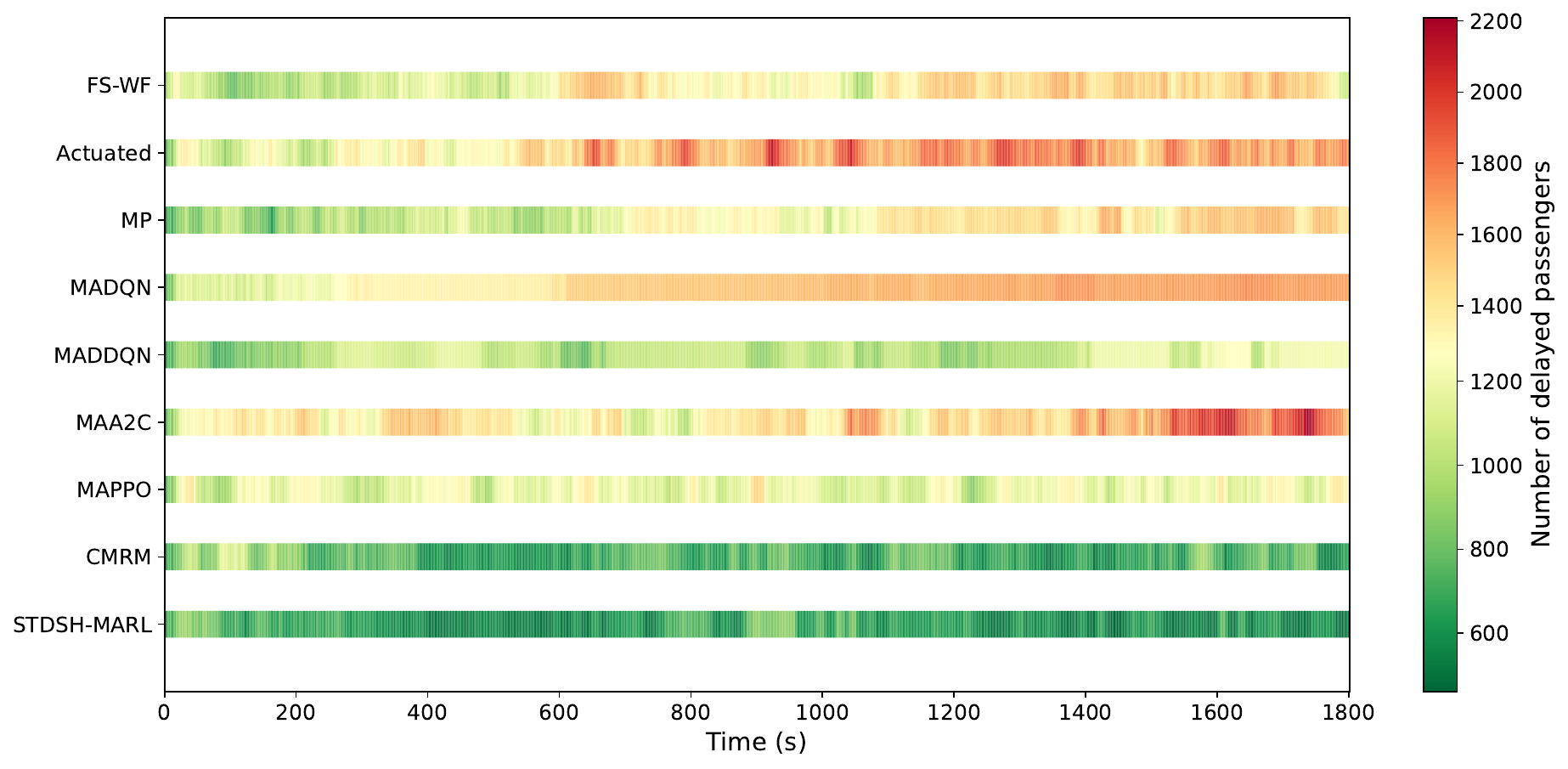}%
  \label{fig:s1a}%
}
\hfill
\subfloat[Vehicle queue length.]{%
  \includegraphics[width=0.95\textwidth]{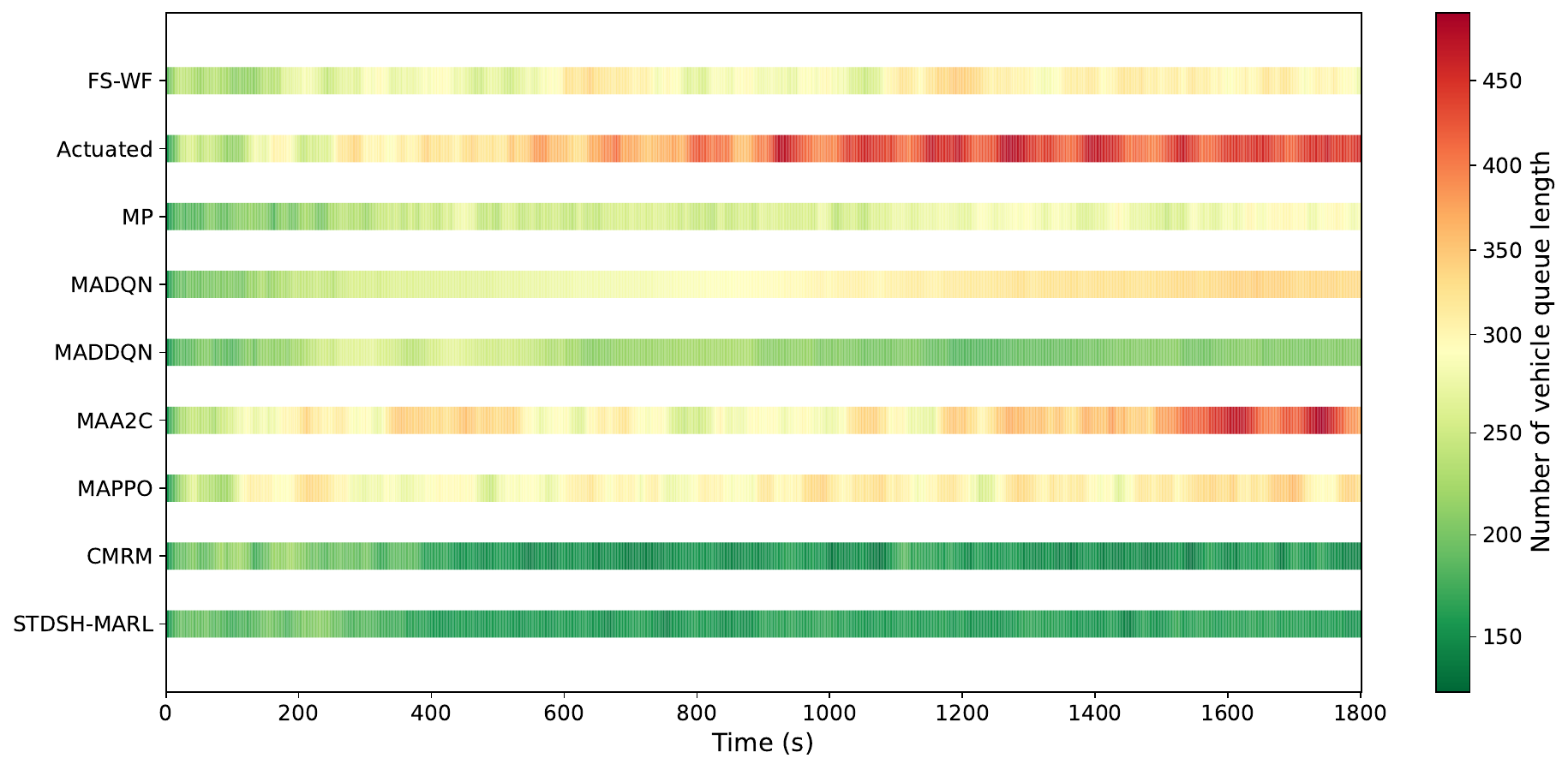}%
  \label{fig:s1b}%
}
\caption{Time-series heatmaps of delayed passengers and vehicle queue lengths over the 1,800-s horizon in Scenario 1, comparing all control methods; color intensity indicates magnitude (cooler = lower, warmer = higher).}
\label{fig:s1}
\end{figure}

\subsection{Scenario 1}
Figure \ref{rewardcurve} shows the reward curve of STDSH-MARL during the training period. The light blue line represents the original cumulative reward, which fluctuates across episodes due to the stochastic nature of the learning process. The dark orange line shows the 10-episode moving average, which smooths these fluctuations and reveals the overall learning trend. 
The cumulative reward improves substantially from approximately -300,000 at the beginning to around -110,000 by Episode 300, indicating that the agent learns a progressively better policy. The largest improvement occurs between roughly Episodes 50 and 130, although the original reward remains highly variable during this period. After about Episode 150–200, the moving-average curve becomes relatively stable and increases more slowly, suggesting that training is approaching convergence with only marginal further improvement.

Table \ref{table_s1} illustrates the performance comparison for scenario 1. This scenario represents a typical off-peak period with relatively low demand. The best metric is labelled in bold and the second-best value is underlined. STDSH-MARL delivers the strongest overall results in Scenario 1, ranking first in three of the four metrics. It achieves the lowest ANP (791.35) and AQL (160.08), outperforming the second-best CMRM by approximately 3.74\% and 3.75\%, respectively. The improvement is much larger relative to the conventional FS-WF strategy, with reductions of 43.55\% in ANP and 46.44\% in AQL. These results indicate that STDSH-MARL is particularly effective at reducing overall traffic pressure and queue-related congestion.

STDSH-MARL also records the lowest AWT (tram) of 88.80 seconds. This represents a 20.71\% reduction compared with FS-WF, the second-best method for this metric, and a 37.79\% reduction compared with CMRM. Compared with the MARL baseline MAPPO, STDSH-MARL reduces ANP, AQL, and AWT (tram) by 46.12\%, 49.29\%, and 27.27\%, respectively, demonstrating the benefit of its proposed spatial–temporal coordination mechanism.

The main trade-off is observed in bus waiting time. STDSH-MARL obtains an AWT (bus) of 371.60 seconds, whereas CMRM achieves the best result of 261.83 seconds and FS-WF achieves the second-best result of 344.13 seconds. Thus, the AWT (bus) of STDSH-MARL is approximately 41.92\% higher than that of CMRM and 7.98\% higher than that of FS-WF. Nevertheless, it still outperforms several learning-based methods, including MAPPO, MADQN, and MADDQN, in terms of AWT (bus). Overall, STDSH-MARL provides the best performance across most metrics, especially for network-level congestion and tram priority, although further improvement may be needed to better balance the waiting times of buses and trams.

The temporal distribution of delayed passengers under Scenario 1 is further illustrated in Figure \ref{fig:s1}. The heatmaps show that STDSH-MARL consistently produces lower intensity delay regions compared with baseline methods, particularly during peak demand intervals. The reduced concentration of high-delay segments visually confirms its superior capability in maintaining corridor-level passenger flow stability.

\begin{table}[htbp]
\centering
\caption{Performance comparison for different models (scenario 1)}
\label{table_s1}
\begin{tabular}{p{3.0cm}p{2.2cm}p{2.2cm}p{2.2cm}p{2.2cm}}
\hline
Model & ANP & AQL & AWT (bus) & AWT (tram) \\
\hline
FS-WF & 1401.86 & 298.88 & \underline{344.13} & \underline{112.00}\\
Actuated & 1671.05 & 390.66 & 350.56 & 169.75\\
MP & 1385.87 & 243.63 & 364.76 & 581.80\\
MADQN & 1680.78 & 282.48 & 944.81 & 1552.67\\
MADDQN & 1123.50 & 192.18 & 687.44 & 484.40\\
MAA2C & 1679.78 & 359.62 & 377.15 & 321.75\\
MAPPO & 1468.63 & 315.68 & 432.78 & 122.10\\
CMRM & \underline{822.11} & \underline{166.32} & \textbf{261.83} & 142.75\\
STDSH-MARL & \textbf{791.35} & \textbf{160.08} & 371.60 & \textbf{88.80}\\
\hline
\end{tabular}
\end{table}

\begin{figure}[htbp]
\centering
\subfloat[Number of delayed passengers.]{%
  \includegraphics[width=0.95\textwidth]{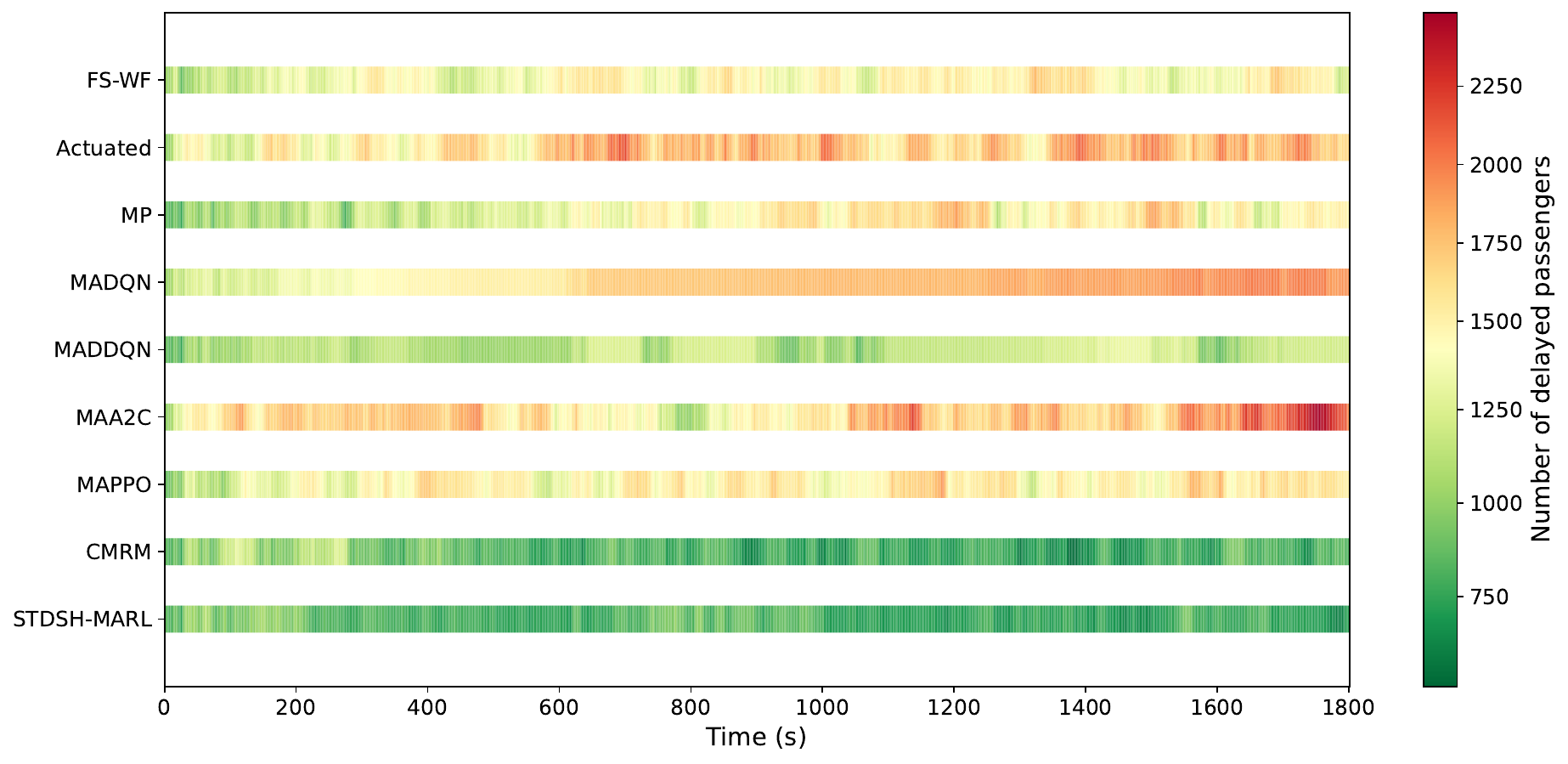}%
  \label{fig:s2a}%
}
\hfill
\subfloat[Vehicle queue length.]{%
  \includegraphics[width=0.95\textwidth]{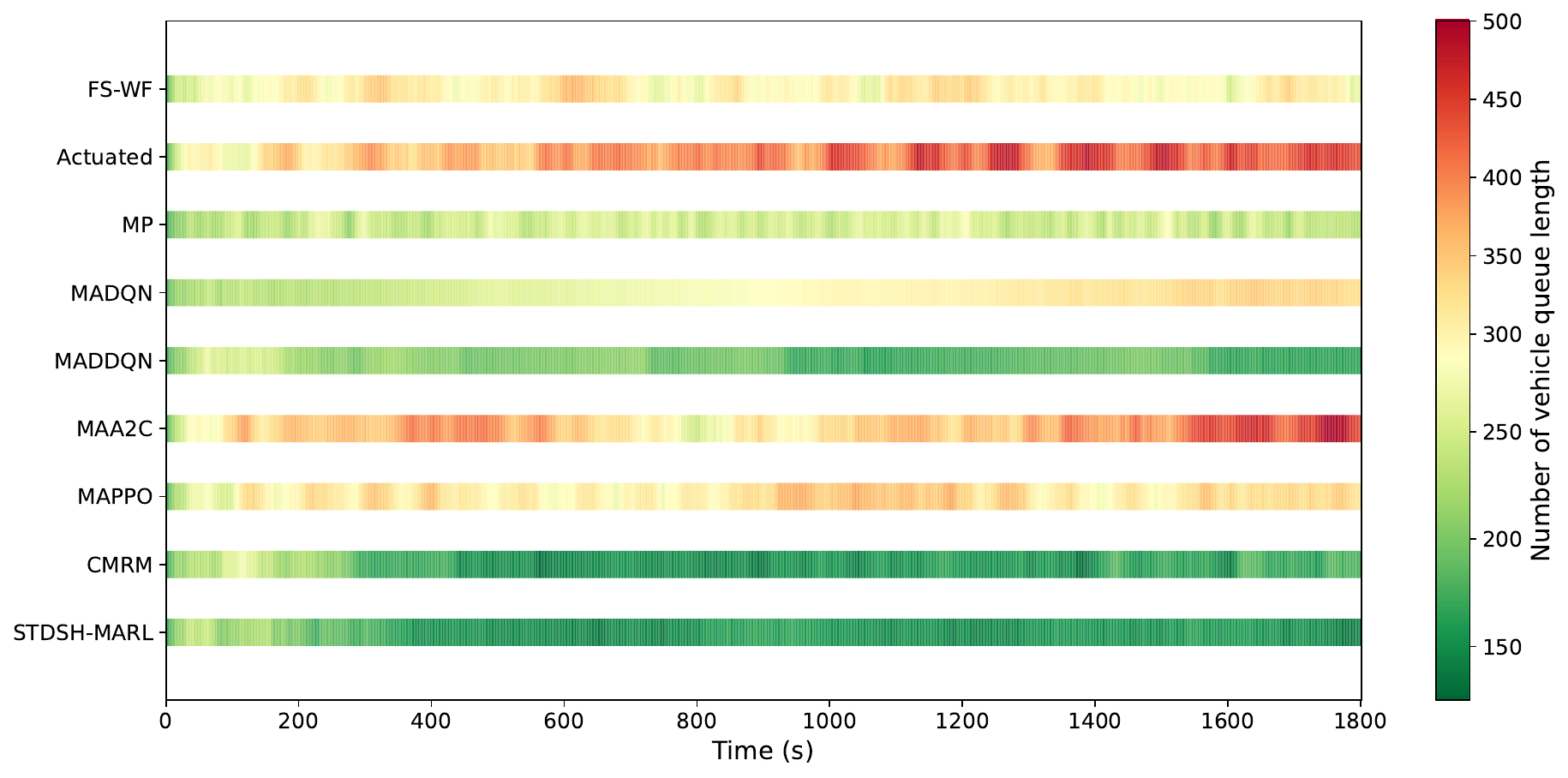}%
  \label{fig:s2b}%
}
\caption{Time-series heatmaps of delayed passengers and vehicle queue lengths over the 1,800-s horizon in Scenario 2, comparing all control methods; color intensity indicates magnitude (cooler = lower, warmer = higher).}
\label{fig:s2}
\end{figure}

\subsection{Scenario 2}
Scenario 2 represents a typical off-peak-to-peak transition period, characterized by gradually increasing traffic demand. Table \ref{table_s2} presents the performance comparison for Scenario 2. 
In this scenario, STDSH-MARL again demonstrates the strongest overall performance, achieving the best results in two of the four metrics. It records the lowest ANP of 623.90, which is 13.10\% lower than the second-best CMRM and 39.49\% lower than MADDQN. This substantial reduction indicates that STDSH-MARL is particularly effective at improving overall network traffic conditions. For AQL, STDSH-MARL achieves 164.12, ranking second and remaining very close to the best result of 161.11 obtained by CMRM. The difference is only about 1.87\%, while STDSH-MARL still reduces AQL by 42.78\% compared with FS-WF and by 45.01\% compared with MAPPO. This suggests that the proposed method maintains highly competitive queue-length performance despite not achieving the absolute minimum.

STDSH-MARL also produces the lowest AWT (tram) of 109.90 seconds. This is 23.84\% lower than CMRM, 11.30\% lower than MAPPO, and 15.14\% lower than FS-WF, demonstrating its effectiveness in prioritising tram movements and reducing tram delays. For AWT (bus), CMRM performs best with 267.73 seconds, while STDSH-MARL ranks second among the learning-based coordination methods with 302.71 seconds. Although this value is 13.07\% higher than that of CMRM, it is still 3.59\% lower than FS-WF and 10.66\% lower than MAPPO.

Overall, STDSH-MARL provides the best balance across the four indicators. It achieves the lowest network pressure and tram waiting time, while remaining close to the best method in average queue length and maintaining competitive bus waiting time. The results indicate that STDSH-MARL is able to improve network-wide traffic efficiency and tram priority without causing a substantial deterioration in bus performance.

The delay patterns over time in Scenario 2 are presented in Figure \ref{fig:s2}. Compared with other DRL baselines, STDSH-MARL exhibits fewer high-delay clusters and smoother temporal transitions. This suggests that the proposed method has stronger adaptability to fluctuating traffic demand and can more effectively prioritize high-occupancy public transport during the transition from off-peak to peak conditions.

\begin{table}[htbp]
\centering
\caption{Performance comparison for different models (scenario 2)}
\label{table_s2}
\begin{tabular}{p{3.0cm}p{2.2cm}p{2.2cm}p{2.2cm}p{2.2cm}}
\hline
Model & ANP & AQL & AWT (bus) & AWT (tram) \\
\hline
FS-WF & 1298.39 & 286.81 & 313.97 & 129.50\\
Actuated & 1542.32 & 378.17 & 315.94 & 172.35\\
MP & 1237.46 & 255.47 & 350.10 & 532.55\\
MADQN & 1501.23 & 289.10 & 843.04 & 1594.54\\
MADDQN & 1031.01 & 214.91 & 502.11 & 640.79\\
MAA2C & 1470.48 & 325.22 & 314.49 & 328.85\\
MAPPO & 1173.20 & 298.47 & 338.84 & \underline{123.90}\\
CMRM & \underline{717.92} & \textbf{161.11} & \textbf{267.73} & 144.30\\
STDSH-MARL & \textbf{623.90} & \underline{164.12} & \underline{302.71} & \textbf{109.90}\\
\hline
\end{tabular}
\end{table}

\begin{figure}[htbp]
\centering
\subfloat[Number of delayed passengers.]{%
  \includegraphics[width=0.95\textwidth]{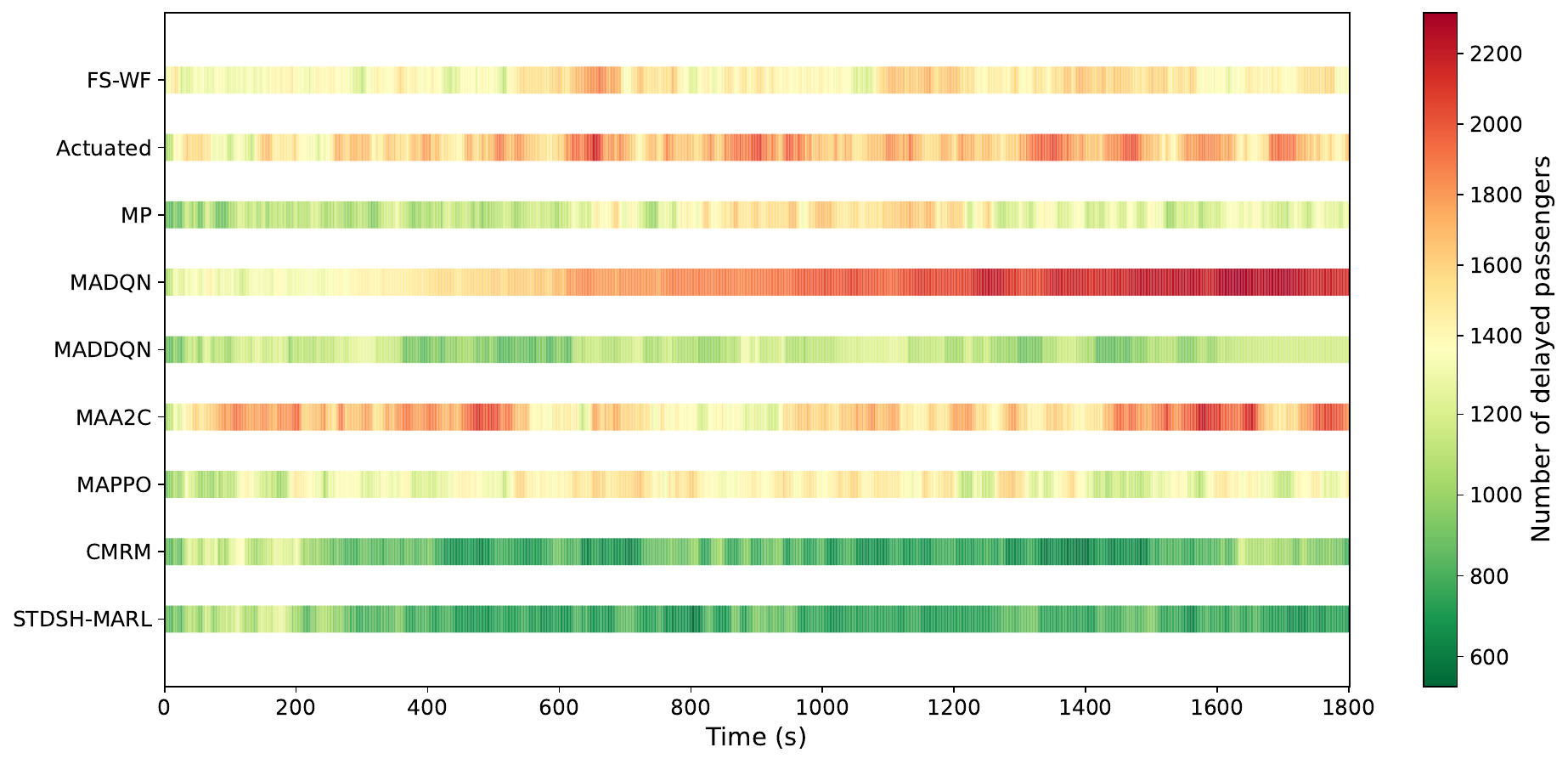}%
  \label{fig:s3a}%
}
\hfill
\subfloat[Vehicle queue length.]{%
  \includegraphics[width=0.95\textwidth]{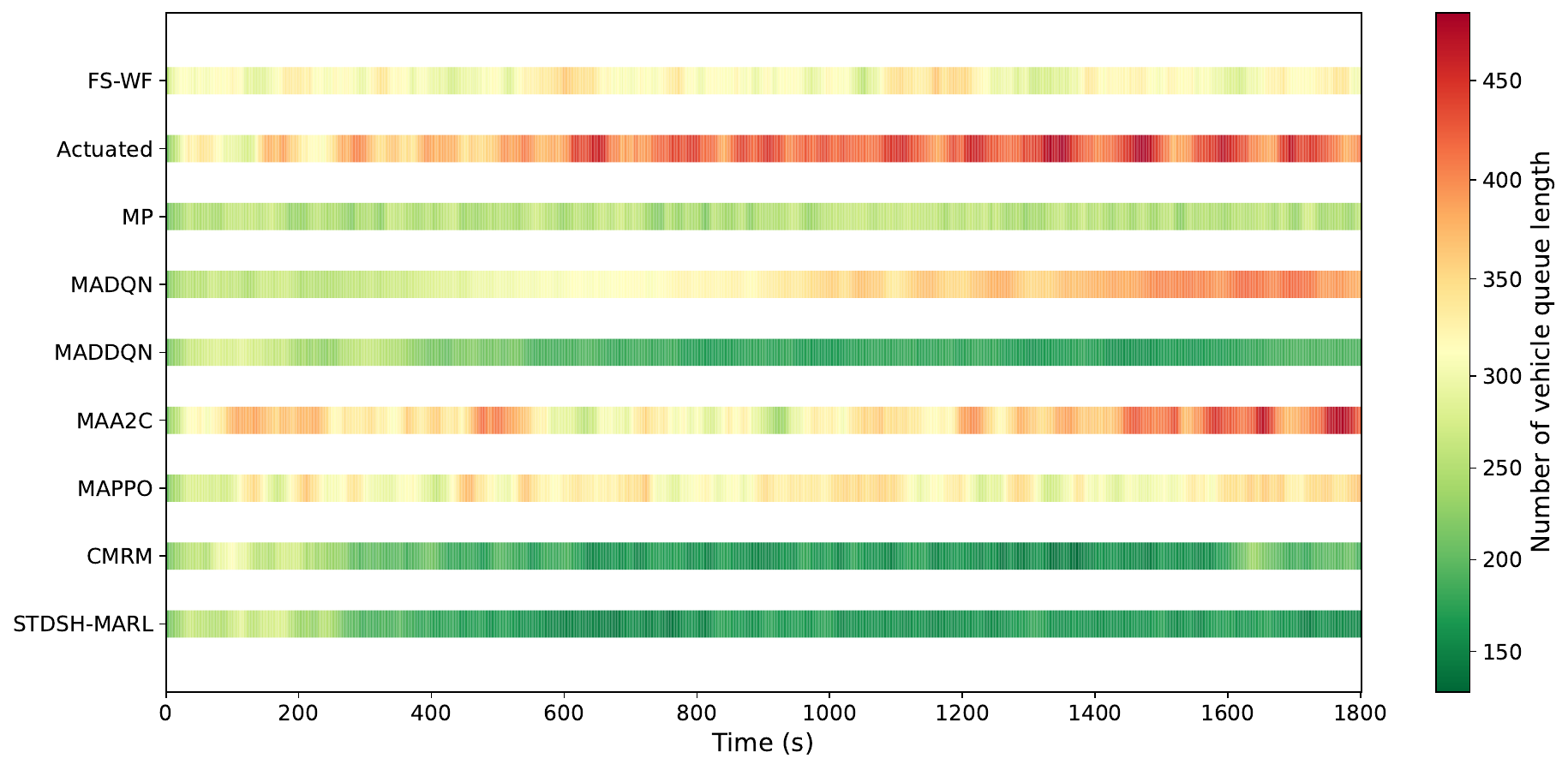}%
  \label{fig:s3b}%
}
\caption{Time-series heatmaps of delayed passengers and vehicle queue lengths over the 1,800-s horizon in Scenario 3, comparing all control methods; color intensity indicates magnitude (cooler = lower, warmer = higher).}
\label{fig:s3}
\end{figure}

\subsection{Scenario 3}
Scenario 3 represents the rush-hour period with higher traffic volumes. Table \ref{table_s3} presents the performance comparison for Scenario 3. STDSH-MARL achieves the strongest overall performance, obtaining the lowest ANP, AQL, and AWT (tram) among all evaluated methods. Its ANP of 760.37 is 9.36\% lower than the second-best CMRM, 33.53\% lower than MADDQN, and 39.73\% lower than MP. Similarly, its AQL of 164.76 is 5.53\% lower than that of CMRM, 15.61\% lower than that of MADDQN, and 32.99\% lower than that of MP. These results indicate that STDSH-MARL is highly effective at alleviating network pressure and reducing congestion under Scenario 3. 

The proposed model also achieves the best AWT (tram) of 90.50 seconds. This represents reductions of 28.85\% compared with FS-WF, 31.88\% compared with MAPPO, and 49.36\% compared with CMRM, demonstrating a clear advantage in tram priority and delay reduction. The largest tram waiting times are observed for MADQN and MADDQN, suggesting that these methods struggle to coordinate tram movements effectively in this scenario.

The main limitation remains bus performance. STDSH-MARL records a bus AWT of 352.23 seconds, which is 47.19\% higher than the best-performing CMRM and 23.87\% higher than MP. Nevertheless, it is slightly better than FS-WF by 0.73\% and improves upon MAPPO by 15.49\%. Overall, STDSH-MARL provides the best network-level and tram-related performance, although the results reveal a trade-off between prioritising tram movements and minimising bus waiting time.

As shown in Figure \ref{fig:s3}, Scenario 3 involves more dynamic demand variations under rush-hour conditions. STDSH-MARL maintains relatively low delay intensity across most time intervals, whereas competing methods exhibit more concentrated delay peaks. This visual comparison further supports the quantitative improvements reported in Table \ref{table_s3}.

\begin{table}[htbp]
\centering
\caption{Performance comparison for different models (scenario 3)}
\label{table_s3}
\begin{tabular}{p{3.0cm}p{2.2cm}p{2.2cm}p{2.2cm}p{2.2cm}}
\hline
Model & ANP & AQL & AWT (bus) & AWT (tram) \\
\hline
FS-WF & 1489.44 & 337.71 & 354.82 & \underline{127.20}\\
Actuated & 1747.40 & 387.99 & 377.09 & 205.85\\
MP & 1261.59 & 245.88 & \underline{284.35} & 423.85\\
MADQN & 1842.39 & 302.48 & 958.39 & 1594.54\\
MADDQN & 1143.86 & 195.24 & 659.35 & 538.63\\
MAA2C & 1642.75 & 349.47 & 362.77 & 317.50\\
MAPPO & 1436.28 & 333.47 & 416.79 & 132.85\\
CMRM & \underline{838.92} & \underline{174.40} & \textbf{239.31} & 178.70\\
STDSH-MARL & \textbf{760.37} & \textbf{164.76} & 352.23 & \textbf{90.50}\\
\hline
\end{tabular}
\end{table}

\begin{figure}[htbp]
\centering
\subfloat[Number of delayed passengers.]{%
  \includegraphics[width=0.95\textwidth]{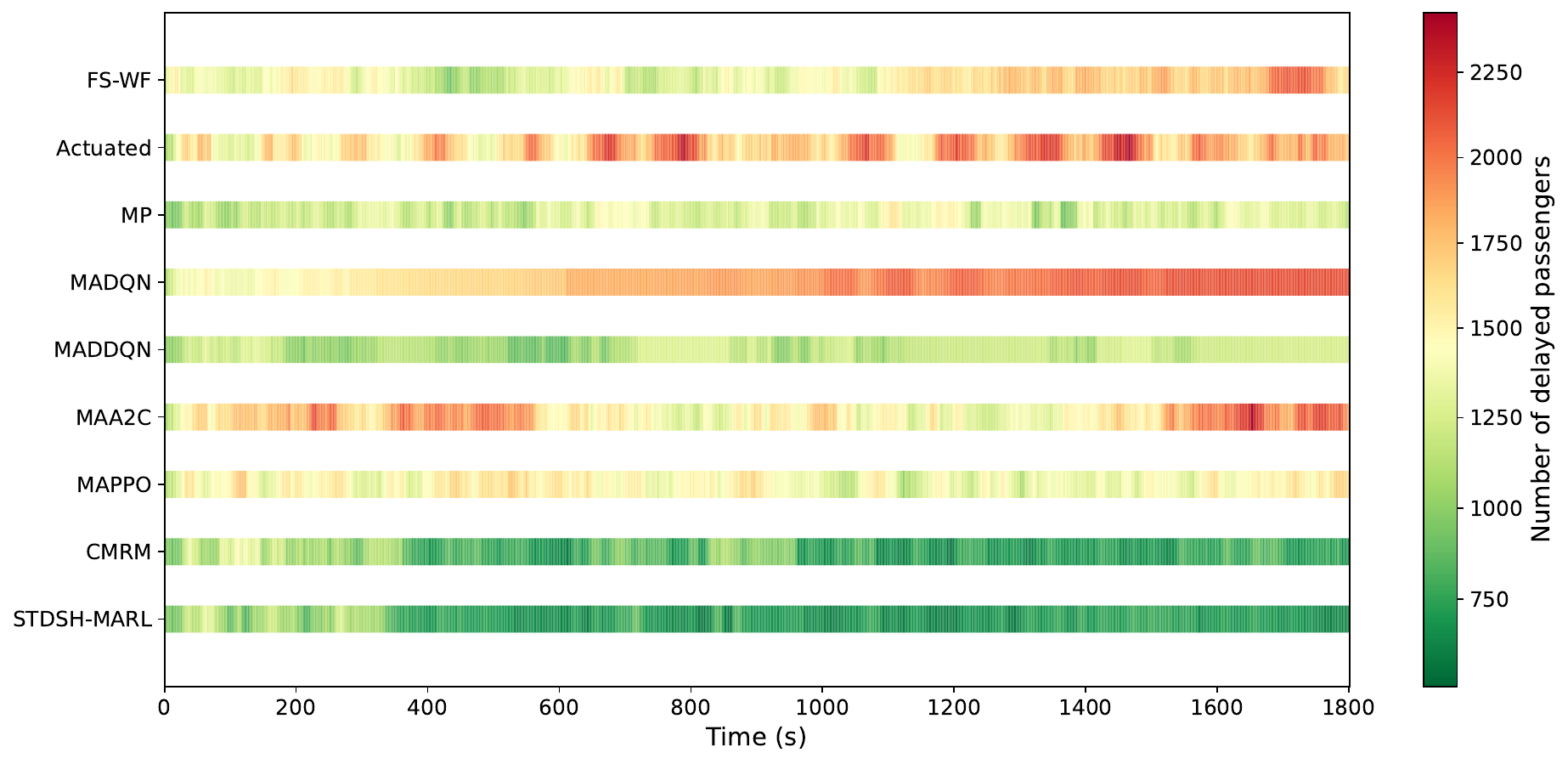}%
  \label{fig:s4a}%
}
\hfill
\subfloat[Vehicle queue length.]{%
  \includegraphics[width=0.95\textwidth]{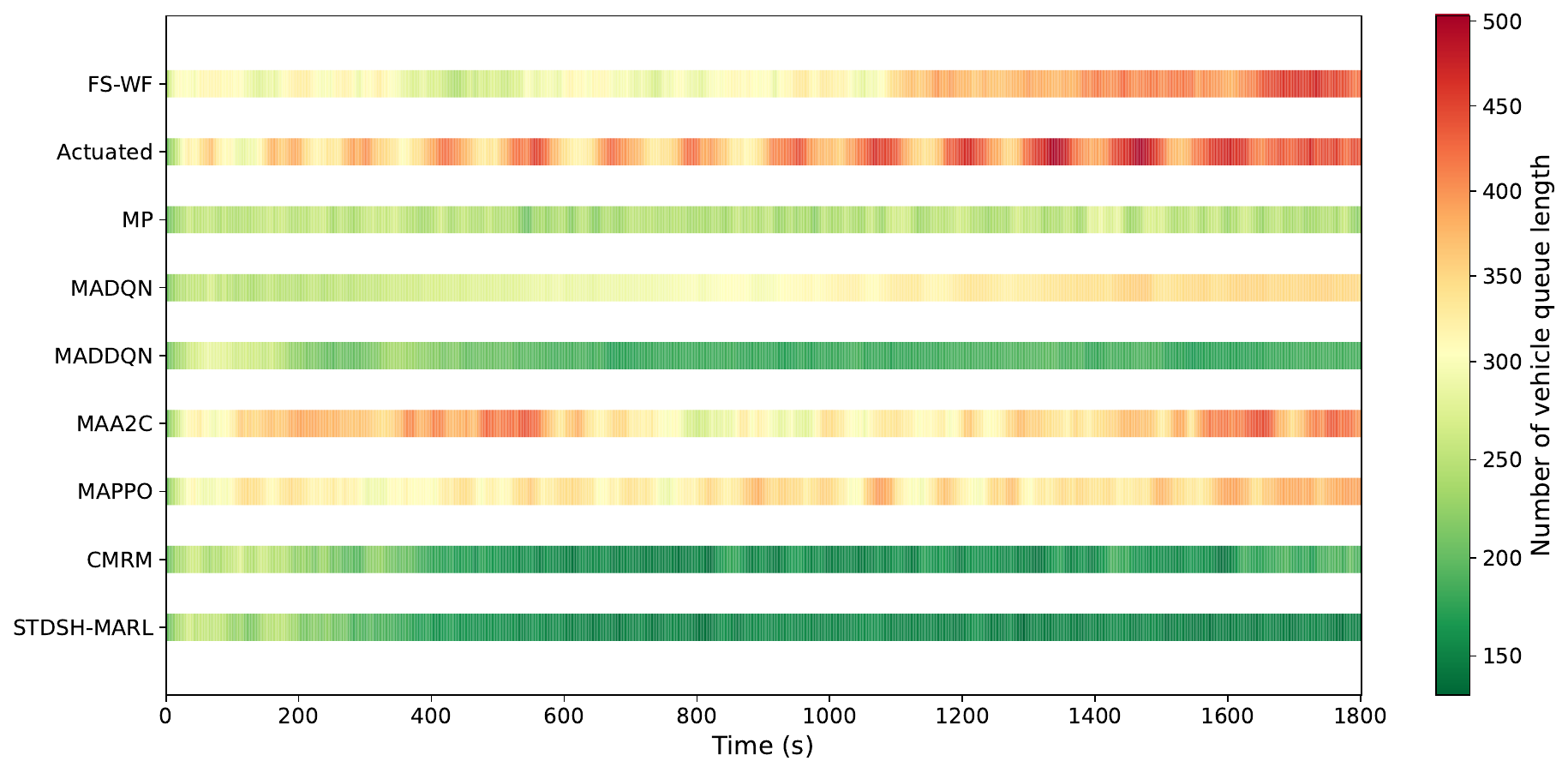}%
  \label{fig:s4b}%
}
\caption{Time-series heatmaps of delayed passengers and vehicle queue lengths over the 1,800-s horizon in Scenario 4, comparing all control methods; color intensity indicates magnitude (cooler = lower, warmer = higher).}
\label{fig:s4}
\end{figure}

\subsection{Scenario 4}
Scenario 4 represents the morning school period, during which traffic demand increases around a specific area. Table \ref{table_s4} presents the performance comparison for Scenario 4. STDSH-MARL achieves the best ANP and AQL among all evaluated methods. Its ANP of 799.44 is 4.09\% lower than the second-best CMRM, 44.96\% lower than FS-WF, and 40.63\% lower than MAPPO. Similarly, its AQL of 176.15 improves upon CMRM by 4.33\% and reduces the queue length by 43.65\% and 44.79\% relative to FS-WF and MAPPO, respectively. These results show that STDSH-MARL remains highly effective in reducing network pressure and congestion under Scenario 4.

For AWT (tram), MAPPO achieves the best result of 87.15 seconds, while STDSH-MARL ranks second with 94.95 seconds. Although the tram waiting time of STDSH-MARL is 8.95\% higher than that of MAPPO, it is still 23.77\% lower than FS-WF and 30.92\% lower than CMRM. This indicates that the proposed method maintains strong tram-priority performance while simultaneously achieving much better network-level efficiency than MAPPO.

For AWT (bus), CMRM performs best with 249.85, followed by MP with 328.72. STDSH-MARL records 350.62, ranking third and remaining almost identical to FS-WF, with a slight reduction of 0.15\%. However, its AWT (bus) is 40.33\% higher than that of CMRM and 6.66\% higher than that of MP. Overall, STDSH-MARL provides the strongest performance in terms of network pressure and queue length and achieves competitive tram waiting time, but its bus waiting-time performance indicates a remaining trade-off between overall traffic efficiency, tram priority, and bus service quality.

The robustness of the proposed framework under higher congestion levels is illustrated in Figure \ref{fig:s4}. STDSH-MARL mitigates extreme delay accumulation more effectively than the competing methods, leading to a more balanced and stable delay distribution across the evaluation horizon.

\begin{table}[htbp]
\centering
\caption{Performance comparison for different models (scenario 4)}
\label{table_s4}
\begin{tabular}{p{3.0cm}p{2.2cm}p{2.2cm}p{2.2cm}p{2.2cm}}
\hline
Model & ANP & AQL & AWT (bus) & AWT (tram) \\
\hline
FS-WF & 1452.45 & 312.61 & 351.15 & 124.56\\
Actuated & 1655.16 & 399.70 & 354.04 & 177.85\\
MP & 1267.16 & 249.59 & \underline{328.72} & 377.85\\
MADQN & 1841.18 & 330.92 & 769.57 & 1594.54\\
MADDQN & 1079.67 & 195.48 & 559.05 & 437.10\\
MAA2C & 1649.37 & 351.65 & 352.28 & 336.00\\
MAPPO & 1346.58 & 319.04 & 382.07 & \textbf{87.15}\\
CMRM & \underline{833.54} & \underline{184.12} & \textbf{249.85} & 137.45\\
STDSH-MARL & \textbf{799.44} & \textbf{176.15} & 350.62 & \underline{94.95}\\
\hline
\end{tabular}
\end{table}

\begin{figure}[htbp]
\centering
\subfloat[Number of delayed passengers.]{%
  \includegraphics[width=0.95\textwidth]{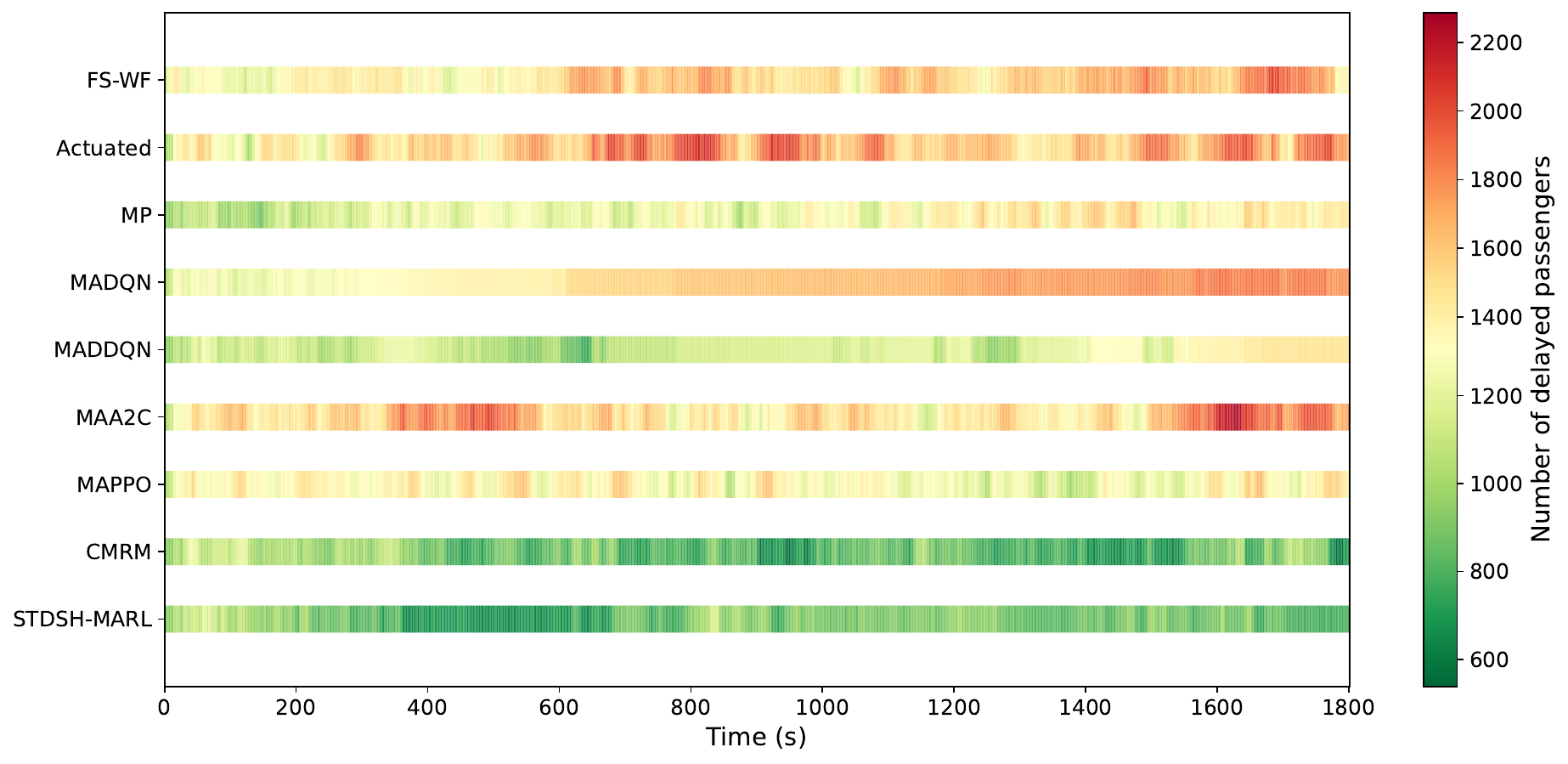}%
  \label{fig:s5a}%
}
\hfill
\subfloat[Vehicle queue length.]{%
  \includegraphics[width=0.95\textwidth]{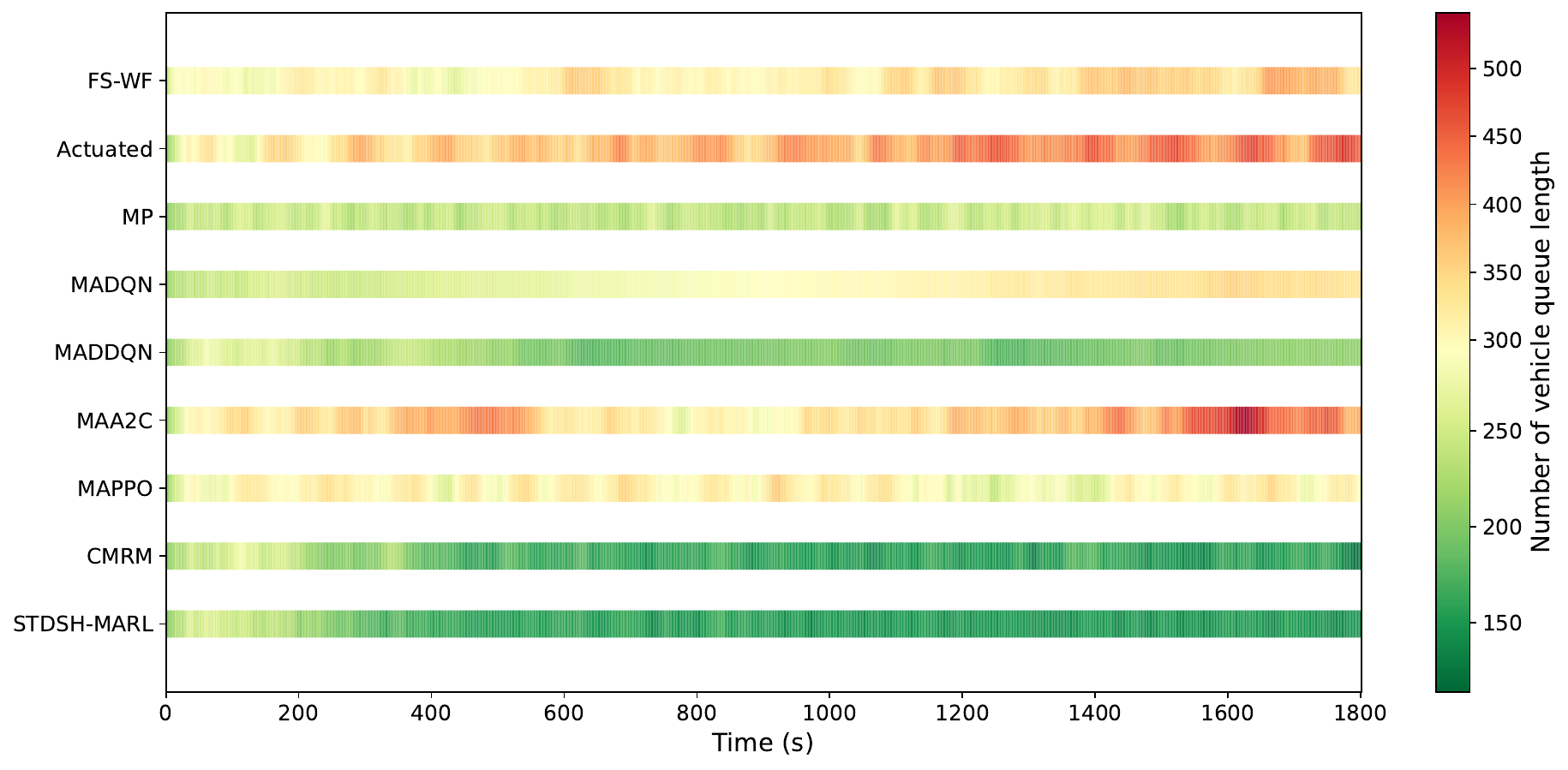}%
  \label{fig:s5b}%
}
\caption{Time-series heatmaps of delayed passengers and vehicle queue lengths over the 1,800-s horizon in Scenario 5, comparing all control methods; color intensity indicates magnitude (cooler = lower, warmer = higher).}
\label{fig:s5}
\end{figure}

\subsection{Scenario 5}
In contrast to scenario 4, this scenario replicates school end periods: scenario 5 describes the afternoon school period with increasing traffic flow away from an area. Table \ref{table_s5} presents the performance comparison of the evaluated methods under this traffic condition.
STDSH-MARL achieves the best network-level performance, recording the lowest ANP of 854.86 and the lowest AQL of 164.14. Its ANP is only 0.23\% lower than that of the second-best CMRM, but it improves upon FS-WF and MAPPO by 43.95\% and 35.18\%, respectively. For AQL, STDSH-MARL outperforms CMRM by 4.79\% and reduces the queue length by 49.23\% compared with FS-WF and 45.71\% compared with MAPPO. These results demonstrate its strong ability to alleviate overall traffic pressure and queue congestion.

For tram AWT, MAPPO achieves the best result of 101.80 seconds, while STDSH-MARL ranks second with a value of 109.60 seconds. Although the tram waiting time of STDSH-MARL is 7.66\% higher than that of MAPPO, it is 21.83\% lower than that of CMRM. These results indicate that STDSH-MARL maintains effective tram priority while achieving substantially better ANP and AQL performance than MAPPO.

The main limitation is AWT (bus). STDSH-MARL records 436.60 seconds, which is 63.28\% higher than the best result achieved by CMRM and 27.58\% higher than the second-best MAA2C. It is also 18.75\% higher than FS-WF and 9.34\% higher than MAPPO. Overall, STDSH-MARL provides the strongest performance in reducing network pressure and queue length and remains highly competitive for tram waiting time, but its relatively high bus waiting time reveals a clear trade-off between network-wide congestion reduction, tram priority, and bus service quality.

The corresponding delay distributions are visualized in Figure~\ref{fig:s5}. Despite the intensified demand, STDSH-MARL suppresses large delay clusters and produces a more uniform temporal delay pattern, indicating its robustness under biased traffic conditions.

\begin{table}[htbp]
\centering
\caption{Performance comparison for different models (scenario 5)}
\label{table_s5}
\begin{tabular}{p{3.0cm}p{2.2cm}p{2.2cm}p{2.2cm}p{2.2cm}}
\hline
Model & ANP & AQL & AWT (bus) & AWT (tram) \\
\hline
FS-WF & 1525.12 & 323.31 & 367.67 & 138.05\\
Actuated & 1632.73 & 385.13 & 365.03 & 166.10\\
MP & 1271.78 & 241.23 & 356.08 & 388.10\\
MADQN & 1567.47 & 292.65 & 890.94 & 1594.54\\
MADDQN & 1165.59 & 206.80 & 575.79 & 656.68\\
MAA2C & 1571.48 & 363.43 & \underline{342.23} & 267.75\\
MAPPO & 1318.77 & 302.34 & 399.30 & \textbf{101.80}\\
CMRM & \underline{856.81} & \underline{172.39} & \textbf{267.39} & 140.20\\
STDSH-MARL & \textbf{854.86} & \textbf{164.14} & 436.60 & \underline{109.60}\\
\hline
\end{tabular}
\end{table}

\subsection{Discussion}
In this section, we first conduct an ablation study to examine the contribution of each component in the proposed STDSH-MARL framework, with particular focus on the STDSH module. Key components include dual-stage hypergraph attention (DSHA), spatial hyperedge (SHE), and temporal hyperedge (THE). Table \ref{table_as} presents the results of the ablation studies over all scenarios. A \checkmark indicates that the component is included in the ablation model, while $\times$ indicates that it is excluded.

The ablation results demonstrate the complementary contributions of DSHA, SHE, and THE within the complete STDSH-MARL framework.
The complete model achieves the lowest average ANP (765.98), AQL (165.85), and tram AWT (98.75), and consistently obtains the best values for these three metrics across all five scenarios. Compared with the variants without DSHA, SHE, and THE, the complete model reduces average ANP by 17.32\%, 48.32\%, and 49.63\%, respectively; average AQL by 14.37\%, 51.06\%, and 51.18\%; and AWT (tram) by 51.08\%, 45.11\%, and 55.15\%. These results demonstrate that jointly modeling spatial and temporal dependencies using the dual-stage attention mechanism improves passenger-level network performance and tram operations. However, the complete model has a higher bus AWT than the ablated variants, indicating a trade-off between the improvements in network-level and tram-related performance and bus waiting time.

Removing THE produces the largest deterioration: average ANP, AQL, and AWT (tram) increase by approximately 98.52\%, 104.82\%, and 122.98\%, respectively. Removing SHE also substantially degrades performance, increasing ANP by 93.49\%, AQL by 104.35\%, and tram AWT by 82.20\%. These results show that both types of hyperedges are essential, while temporal hyperedges appear particularly important for reducing tram waiting time, likely because they capture traffic evolution and signal-state dependencies over time.

Removing DSHA while retaining both spatial and temporal hyperedges increases ANP by 20.95\%, AQL by 16.79\%, and tram AWT by 104.42\%. The smaller deterioration in ANP and AQL suggests that the underlying hypergraph structure alone can still capture useful spatial–temporal relationships. However, the substantial increase in tram AWT shows that DSHA is important for adaptively weighting relevant nodes and hyperedges when coordinating tram movements.

A trade-off is observed for bus performance. The complete model has an average AWT (bus) of 362.75, compared with 339.96 without THE, 340.98 without SHE, and 242.97 without DSHA. Thus, the improvements in network efficiency and tram priority are achieved at the cost of increased bus waiting time. Overall, the ablation study confirms the complementary contributions of DSHA, SHE, and THE, while also suggesting that the reward design or priority-balancing mechanism could be further refined to improve bus service without sacrificing the substantial gains in congestion reduction and tram performance. Based on the ANP values, the order of importance of each component can be ranked as: THE > SHE > DSHA. The most important component is THE.

The ablation results indicate that the spatial and temporal hyperedges are complementary rather than independently additive. MAPPO, which uses no hypergraph representation, achieves an average ANP of 1348.69, while removing THE or SHE increases ANP to 1520.61 and 1482.09, respectively, suggesting that an incomplete hypergraph may provide a less informative representation than no hypergraph. In contrast, retaining both SHE and THE without DSHA reduces ANP to 926.49, and the complete model further reduces it to 765.98. These results show that the main performance gain arises from jointly modeling spatial and temporal dependencies, with DSHA further enhancing their interaction, rather than from either hyperedge type in isolation.

\begin{table}[htbp]
\centering
\caption{Ablation study performance of each component across all scenarios}
\label{table_as}
\begin{threeparttable}
\begin{tabular}{p{1.1cm}p{1.1cm}p{1.1cm}p{1.6cm}p{1.6cm}p{1.6cm}p{1.9cm}p{2.0cm}}
\hline
DSHA & SHE & THE & Scenario & ANP & AQL & AWT (bus) & AWT (tram) \\
\hline
\checkmark & \checkmark & $\times$ & 1 & 1511.44 & 324.95 & 357.38 & 239.70\\
\checkmark & \checkmark & $\times$ & 2 & 1446.86 & 334.88 & 297.68 & 226.30\\
\checkmark & \checkmark & $\times$ & 3 & 1584.63 & 347.83 & 354.56 & 189.65\\
\checkmark & \checkmark & $\times$ & 4 & 1572.83 & 355.27 & 353.03 & 205.25\\
\checkmark & \checkmark & $\times$ & 5 & 1487.27 & 335.51 & 337.15 & 240.05\\
\checkmark & \checkmark & $\times$ & Average & 1520.61 & 339.69 & \underline{339.96} & 220.19\\
\hline
\checkmark & $\times$ & \checkmark & 1 & 1470.67 & 327.57 & 335.05 & 226.05\\
\checkmark & $\times$ & \checkmark & 2 & 1429.36 & 326.81 & 350.34 & 165.75\\
\checkmark & $\times$ & \checkmark & 3 & 1489.68 & 338.28 & 345.87 & 151.45\\
\checkmark & $\times$ & \checkmark & 4 & 1586.85 & 364.93 & 372.60 & 149.90\\
\checkmark & $\times$ & \checkmark & 5 & 1433.89 & 336.93 & 301.04 & 206.45\\
\checkmark & $\times$ & \checkmark & Average & 1482.09 & 338.91 & 340.98 & \underline{179.92}\\
\hline
$\times$ & \checkmark & \checkmark & 1 & 890.38 & 172.83 & 301.18 & 149.35\\
$\times$ & \checkmark & \checkmark & 2 & 848.84 & 185.58 & 218.15 & 194.55\\
$\times$ & \checkmark & \checkmark & 3 & 928.40 & 198.21 & 214.88 & 195.45\\
$\times$ & \checkmark & \checkmark & 4 & 1043.88 & 212.23 & 249.89 & 272.30\\
$\times$ & \checkmark & \checkmark & 5 & 920.93 & 199.61 & 230.73 & 197.65\\
$\times$ & \checkmark & \checkmark & Average & \underline{926.49} & \underline{193.69} & \textbf{242.97} & 201.86\\
\hline
\checkmark & \checkmark & \checkmark & 1 & 791.35 & 160.08 & 371.60 & 88.80\\
\checkmark & \checkmark & \checkmark & 2 & 623.90 & 164.12 & 302.71 & 109.90\\
\checkmark & \checkmark & \checkmark & 3 & 760.37 & 164.76 & 352.23 & 90.50\\
\checkmark & \checkmark & \checkmark & 4 & 799.44 & 176.15 & 350.62 & 94.95\\
\checkmark & \checkmark & \checkmark & 5 & 854.86 & 164.14 & 436.60 & 109.60\\
\checkmark & \checkmark & \checkmark & Average & \textbf{765.98} & \textbf{165.85} & 362.75 & \textbf{98.75}\\
\hline
\end{tabular}
\begin{tablenotes}
\item[]\textit{Notes:} “DSHA” denotes dual-stage hypergraph attention; “SHE” denotes spatial hyperedge; and “THE” denotes temporal hyperedge. A \checkmark indicates that the component is included in the ablation model, while $\times$ indicates that it is excluded.
\end{tablenotes}
\end{threeparttable}
\end{table}

\begin{table}[htbp]
\centering
\caption{Statistical significance testing of number of passengers experiencing delay between the proposed STDSH-MARL and some baseline methods (on Scenario 1).}
\label{tab:ttest}
\begin{tabular}{p{5.0cm}p{2.6cm}p{2.6cm}p{2.6cm}}
\hline
Comparison & Proposed Mean & Baseline Mean & $p$-value \\
\hline
STDSH-MARL vs MADQN  & 791.35 & 1680.78 & 0.0 \\
STDSH-MARL vs MADDQN  & 791.35 & 1123.50 & 0.0 \\
STDSH-MARL vs MAA2C & 791.35 & 1679.78 & 0.0 \\
STDSH-MARL vs MAPPO  & 791.35 & 1468.63 & 0.0 \\
STDSH-MARL vs MP  & 791.35 & 1385.87 & 0.0 \\
STDSH-MARL vs CMRM  & 791.35 & 822.11 & $4.01 \times 10^{-125}$ \\
\hline
\end{tabular}

\vspace{1mm}
\begin{minipage}{1\textwidth}
\footnotesize
\raggedright
Note: One-sided Welch's $t$-tests are conducted with the alternative hypothesis 
$H_1: \mu_{\text{Proposed STDSH-MARL}} < \mu_{\text{Baseline}}$, 
against the null hypothesis 
$H_0: \mu_{\text{Proposed STDSH-MARL}} \geq \mu_{\text{Baseline}}$.
\end{minipage}
\end{table}

Table \ref{tab:ttest} reports the statistical significance test results for the number of passengers experiencing delay under Scenario 1. STDSH-MARL achieves the lowest mean value, 791.35, outperforming all evaluated baseline methods, including MP, MADQN, MADDQN, MAA2C, MAPPO, and CMRM. The improvements indicate that the proposed spatio-temporal dual-stage hypergraph representation offers substantial advantages over conventional value-based and actor--critic learning approaches. The one-sided Welch's t-tests yield extremely small $p$-values; most are reported as 0.00 at the displayed numerical precision, while the comparison with CMRM produces a $p$-value of $4.01 \times 10^{-125}$. All values are well below the conventional significance threshold of 0.001, confirming that the reductions achieved by STDSH-MARL are statistically significant rather than attributable to random variation. Overall, Table \ref{tab:ttest} provides strong statistical evidence that the proposed method improves the human-centric objective by significantly reducing passenger-level delay compared with the selected baselines.

\begin{figure}[htbp]
\begin{center}
\includegraphics[width=0.66\textwidth]{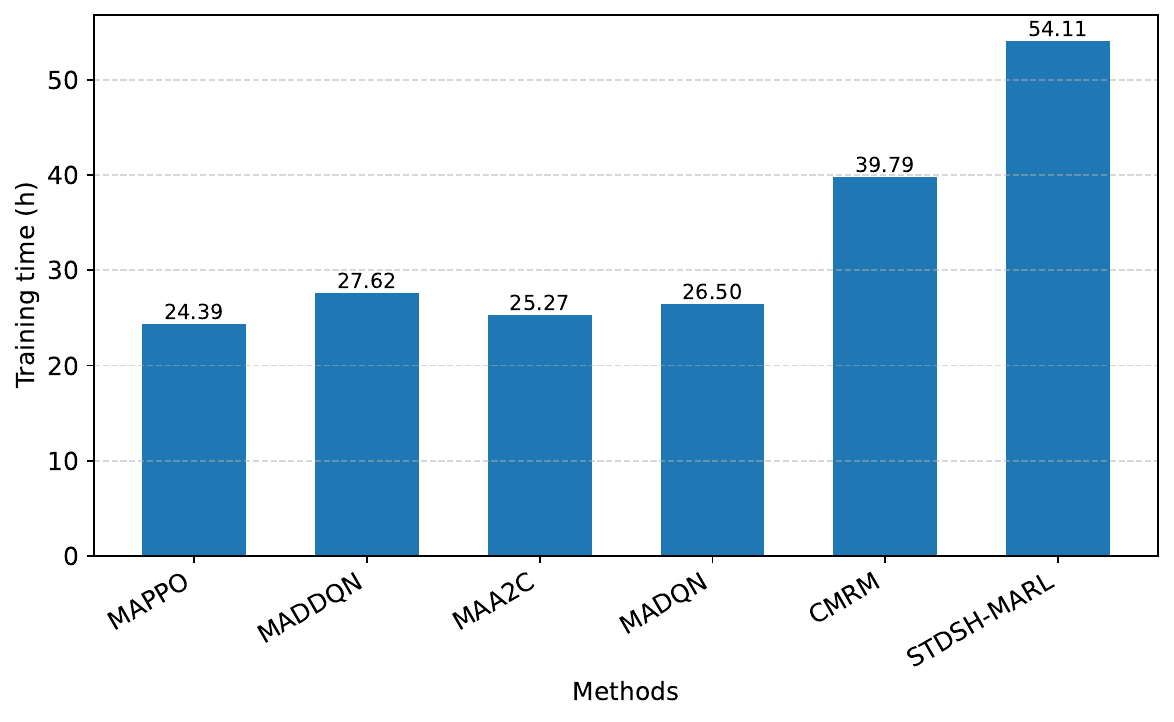}
\caption{Training time comparison among different control methods.}
\label{fig:training_time}
\end{center}
\end{figure}

Figure \ref{fig:training_time} compares the computational training costs of the evaluated control methods. The baseline approaches exhibit broadly comparable training times, ranging from 24.39 h for MAPPO to 39.79 h for CMRM, suggesting similar computational demands among conventional DRL-based methods. In contrast, STDSH-MARL requires 54.11 h of training, exceeding all baseline methods. This additional computational burden can be attributed to the more sophisticated spatio-temporal dual-stage hypergraph representation and multi-agent coordination mechanism incorporated into the proposed framework. Nevertheless, the superior passenger-level performance observed across the evaluated scenarios indicates a clear trade-off between increased training cost and improved control effectiveness.

The current hypergraph construction is relatively coarse. Each spatial hyperedge includes all intersections at a given time step, while each temporal hyperedge includes all observations of one intersection within the historical window. This design provides a simple and unified way to represent corridor-level spatial and temporal information without manually defining many hyperedges. However, it does not explicitly capture finer relationships based on physical adjacency, traffic-flow direction, route connectivity, travel time, or traffic-state similarity. Although the attention mechanism can assign lower weights to less relevant nodes, it is still constrained by the predefined hyperedge memberships. As a result, broad spatial hyperedges may weaken strong local interactions, while full-window temporal hyperedges may not distinguish short-term from long-term dependencies. These limitations may become more significant in larger or more complex networks. Future work will investigate sparse, multi-scale, and dynamically constructed hyperedges based on network structure and changing traffic conditions.

\section{Conclusions and future work}
\label{conclusion}
This paper presented STDSH-MARL, a novel Spatio-Temporal Dual-Stage Hypergraph-based Multi-Agent Reinforcement Learning framework for human-centric traffic signal control in multimodal corridor networks. By integrating a dual-stage hypergraph attention mechanism that captures both spatial and temporal dependencies across intersections, the model enables more effective and adaptive traffic signal coordination. The proposed hybrid action space jointly selects signal phase and green duration, improving control flexibility and responsiveness, particularly for high-occupancy public transport. Comprehensive experiments across five distinct traffic scenarios demonstrate the strong overall performance of STDSH-MARL compared with a wide range of baseline methods, including value-based, policy-based, and graph-based DRL frameworks. The model achieved substantial and relatively consistent reductions in tram waiting time, while its effects on bus waiting time were more variable across traffic scenarios. These findings indicate that prioritizing high-occupancy public transport involves trade-offs among overall network efficiency, tram priority, and bus service quality, rather than uniformly improving all public-transport performance measures. Nevertheless, the overall results demonstrate the potential of STDSH-MARL to support public-transport priority while balancing multimodal traffic performance. Ablation studies and discussions further confirmed that temporal hyperedges within the dual-stage attention mechanism played a particularly crucial role in driving the observed performance improvements.

While this study evaluates STDSH-MARL on a six-intersection corridor, extending the framework to larger, branched, or irregular networks may require topology-adaptive hypergraph construction based on network connectivity and traffic interactions, together with sparse or hierarchical representations to improve scalability. Future work will therefore evaluate STDSH-MARL on larger-scale networks using calibrated simulations, with particular attention to training time, memory usage, and network-wide coordination. Controlled encoder-level comparisons with graph- and Transformer-based alternatives under otherwise matched MARL settings will also be conducted to better isolate the contribution of the proposed hypergraph representation. In addition, we will assess performance under realistic disruptions, including incidents, abrupt demand changes, and noisy detector data, and extend the objectives to incorporate a broader range of multimodal priority measures. Finally, the framework will be further developed toward explainable deep reinforcement learning, providing transparent and auditable explanations for signal-control decisions to enhance operator trust and facilitate practical deployment. 

\section*{CRediT authorship contribution statement}
\textbf{Xiaocai Zhang: }Conceptualization, Data curation, Methodology, Software, Writing - original draft.
\textbf{Neema Nassir: }Conceptualization, Project administration, Writing - review \& editing.
\textbf{Milad Haghani: }Conceptualization, Writing - review \& editing.

\section*{Declaration of competing interest}
The authors declare that they have no known competing financial interests or personal relationships that could have appeared to influence the work reported in this paper.

\section*{Acknowledgment}
This work was funded by ARC LP200301389, Kapsch TrafficCom Australia, RACQ, and iMOVE CRC, the Cooperative Research Centres program, an Australian Government initiative.

\bibliography{mybibfile}

@article{chu2019multi,
  title={Multi-agent deep reinforcement learning for large-scale traffic signal control},
  author={Chu, Tianshu and Wang, Jie and Codec{\`a}, Lara and Li, Zhaojian},
  journal={IEEE Transactions on Intelligent Transportation Systems},
  volume={21},
  number={3},
  pages={1086--1095},
  year={2019},
  publisher={IEEE}
}

@article{nie2025cmrm,
  title={CMRM: Collaborative Multi-agent Reinforcement Learning for Multi-objective Traffic Signal Control},
  author={Nie, Lei and Qi, Dandan and Liu, Bingyi and Li, Peng and Bao, Haizhou and He, Heng},
  journal={IEEE Transactions on Consumer Electronics},
  year={2025},
  publisher={IEEE}
}

@article{zhang2025towards,
  title={Towards fair lights: A multi-agent masked deep reinforcement learning for efficient corridor-level traffic signal control},
  author={Zhang, Xiaocai and Chan, Lok Sang and Nassir, Neema and Sarvi, Majid},
  journal={Communications in Transportation Research},
  volume={5},
  pages={100203},
  year={2025},
  publisher={Elsevier}
}

@inproceedings{wei2019presslight,
  title={Presslight: Learning max pressure control to coordinate traffic signals in arterial network},
  author={Wei, Hua and Chen, Chacha and Zheng, Guanjie and Wu, Kan and Gayah, Vikash and Xu, Kai and Li, Zhenhui},
  booktitle={Proceedings of the 25th ACM SIGKDD International Conference on Knowledge Discovery \& Data Mining},
  pages={1290--1298},
  year={2019}
}

@article{hunt1982scoot,
  title={The SCOOT on-line traffic signal optimisation technique},
  author={Hunt, PB and Robertson, DI and Bretherton, RD and Royle, M Cr},
  journal={Traffic Engineering \& Control},
  volume={23},
  number={4},
  year={1982}
}

@inproceedings{stevanovic2009scoot,
  title={Scoot and scats: A closer look into their operations},
  author={Stevanovic, Aleksandar and Kergaye, Cameron and Martin, Peter T},
  booktitle={88th Annual Meeting of the Transportation Research Board. Washington DC},
  year={2009}
}

@inproceedings{mccannfeet,
  title={FEET-FIRST: USING SCATS TO IMPROVE PEDESTRIAN PROVISION IN DUBLIN},
  author={McCann, Barry and Curley, Robert},
  booktitle={The 20th JCT Traffic Signal Symposium \& Exhibition},
  year={2015}
}

@inproceedings{wei2018intellilight,
  title={Intellilight: A reinforcement learning approach for intelligent traffic light control},
  author={Wei, Hua and Zheng, Guanjie and Yao, Huaxiu and Li, Zhenhui},
  booktitle={Proceedings of the 24th ACM SIGKDD International Conference on Knowledge Discovery \& Data Mining},
  pages={2496--2505},
  year={2018}
}

@article{yazdani2023intelligent,
  title={Intelligent vehicle pedestrian light (IVPL): A deep reinforcement learning approach for traffic signal control},
  author={Yazdani, Mobin and Sarvi, Majid and Bagloee, Saeed Asadi and Nassir, Neema and Price, Jeff and Parineh, Hossein},
  journal={Transportation Research Part C: Emerging Technologies},
  volume={149},
  pages={103991},
  year={2023},
  publisher={Elsevier}
}

@inproceedings{prabuchandran2014multi,
  title={Multi-agent reinforcement learning for traffic signal control},
  author={Prabuchandran, KJ and AN, Hemanth Kumar and Bhatnagar, Shalabh},
  booktitle={17th International IEEE Conference on Intelligent Transportation Systems (ITSC)},
  pages={2529--2534},
  year={2014},
  organization={IEEE}
}

@article{gong2019decentralized,
  title={Decentralized network level adaptive signal control by multi-agent deep reinforcement learning},
  author={Gong, Yaobang and Abdel-Aty, Mohamed and Cai, Qing and Rahman, Md Sharikur},
  journal={Transportation Research Interdisciplinary Perspectives},
  volume={1},
  pages={100020},
  year={2019},
  publisher={Elsevier}
}

@article{ge2019cooperative,
  title={Cooperative deep Q-learning with Q-value transfer for multi-intersection signal control},
  author={Ge, Hongwei and Song, Yumei and Wu, Chunguo and Ren, Jiankang and Tan, Guozhen},
  journal={IEEE Access},
  volume={7},
  pages={40797--40809},
  year={2019},
  publisher={IEEE}
}

@article{kumar2020fuzzy,
  title={Fuzzy inference enabled deep reinforcement learning-based traffic light control for intelligent transportation system},
  author={Kumar, Neetesh and Rahman, Syed Shameerur and Dhakad, Navin},
  journal={IEEE Transactions on Intelligent Transportation Systems},
  volume={22},
  number={8},
  pages={4919--4928},
  year={2020},
  publisher={IEEE}
}

@inproceedings{wang2024fuel,
  title={Fuel-saving route planning with data-driven and learning-based approaches: A systematic solution for harbor tugs},
  author={Wang, Shengming and Zhang, Xiaocai and Li, Jing and Wei, Xiaoyang and Lau, Hoong Chuin and Dai, Bing Tian and HUANG, Binbin Huang and Xiao, Zhe and Fu, Xiuju and Qin, Zheng},
  booktitle={Proceedings of the International Joint Conference on Artificial Intelligence},
  year={2024},
  pages={7483--7490}
}

@article{su2023emvlight,
  title={EMVLight: A multi-agent reinforcement learning framework for an emergency vehicle decentralized routing and traffic signal control system},
  author={Su, Haoran and Zhong, Yaofeng D and Chow, Joseph YJ and Dey, Biswadip and Jin, Li},
  journal={Transportation Research Part C: Emerging Technologies},
  volume={146},
  pages={103955},
  year={2023},
  publisher={Elsevier}
}

@article{yang2021ihg,
  title={IHG-MA: Inductive heterogeneous graph multi-agent reinforcement learning for multi-intersection traffic signal control},
  author={Yang, Shantian and Yang, Bo and Kang, Zhongfeng and Deng, Lihui},
  journal={Neural Networks},
  volume={139},
  pages={265--277},
  year={2021},
  publisher={Elsevier}
}

@article{wu2022distributed,
  title={Distributed agent-based deep reinforcement learning for large scale traffic signal control},
  author={Wu, Qiang and Wu, Jianqing and Shen, Jun and Du, Bo and Telikani, Akbar and Fahmideh, Mahdi and Liang, Chao},
  journal={Knowledge-Based Systems},
  volume={241},
  pages={108304},
  year={2022},
  publisher={Elsevier}
}

@article{zhu2023multi,
  title={Multi-agent broad reinforcement learning for intelligent traffic light control},
  author={Zhu, Ruijie and Li, Lulu and Wu, Shuning and Lv, Pei and Li, Yafei and Xu, Mingliang},
  journal={Information Sciences},
  volume={619},
  pages={509--525},
  year={2023},
  publisher={Elsevier}
}

@article{yang2023hierarchical,
  title={Hierarchical graph multi-agent reinforcement learning for traffic signal control},
  author={Yang, Shantian},
  journal={Information Sciences},
  volume={634},
  pages={55--72},
  year={2023},
  publisher={Elsevier}
}

@article{liu2023multiple,
  title={Multiple intersections traffic signal control based on cooperative multi-agent reinforcement learning},
  author={Liu, Junxiu and Qin, Sheng and Su, Min and Luo, Yuling and Wang, Yanhu and Yang, Su},
  journal={Information Sciences},
  volume={647},
  pages={119484},
  year={2023},
  publisher={Elsevier}
}

@article{bie2024multi,
  title={Multi-agent Deep Reinforcement Learning collaborative Traffic Signal Control method considering intersection heterogeneity},
  author={Bie, Yiming and Ji, Yuting and Ma, Dongfang},
  journal={Transportation Research Part C: Emerging Technologies},
  volume={164},
  pages={104663},
  year={2024},
  publisher={Elsevier}
}

@article{wang2024large,
  title={A large-scale traffic signal control algorithm based on multi-layer graph deep reinforcement learning},
  author={Wang, Tao and Zhu, Zhipeng and Zhang, Jing and Tian, Junfang and Zhang, Wenyi},
  journal={Transportation Research Part C: Emerging Technologies},
  volume={162},
  pages={104582},
  year={2024},
  publisher={Elsevier}
}

@article{song2024cooperative,
  title={Cooperative traffic signal control through a counterfactual multi-agent deep actor critic approach},
  author={Song, Xiang Ben and Zhou, Bin and Ma, Dongfang},
  journal={Transportation Research Part C: Emerging Technologies},
  volume={160},
  pages={104528},
  year={2024},
  publisher={Elsevier}
}

@article{ren2024two,
  title={Two-layer coordinated reinforcement learning for traffic signal control in traffic network},
  author={Ren, Fuyue and Dong, Wei and Zhao, Xiaodong and Zhang, Fan and Kong, Yaguang and Yang, Qiang},
  journal={Expert Systems with Applications},
  volume={235},
  pages={121111},
  year={2024},
  publisher={Elsevier}
}

@inproceedings{liu2023gplight,
  title={GPLight: Grouped Multi-agent Reinforcement Learning for Large-scale Traffic Signal Control.},
  author={Liu, Yilin and Luo, Guiyang and Yuan, Quan and Li, Jinglin and Jin, Lei and Chen, Bo and Pan, Rui},
  booktitle={IJCAI},
  pages={199--207},
  year={2023}
}

@article{abdoos2020cooperative,
  title={A cooperative multiagent system for traffic signal control using game theory and reinforcement learning},
  author={Abdoos, Monireh},
  journal={IEEE Intelligent Transportation Systems Magazine},
  volume={13},
  number={4},
  pages={6--16},
  year={2020},
  publisher={IEEE}
}

@article{jiang2021distributed,
  title={A distributed multi-agent reinforcement learning with graph decomposition approach for large-scale adaptive traffic signal control},
  author={Jiang, Shan and Huang, Yufei and Jafari, Mohsen and Jalayer, Mohammad},
  journal={IEEE Transactions on Intelligent Transportation Systems},
  volume={23},
  number={9},
  pages={14689--14701},
  year={2021},
  publisher={IEEE}
}

@article{zhang2022distributed,
  title={Distributed signal control of arterial corridors using multi-agent deep reinforcement learning},
  author={Zhang, Weibin and Yan, Chen and Li, Xiaofeng and Fang, Liangliang and Wu, Yao-Jan and Li, Jun},
  journal={IEEE Transactions on Intelligent Transportation Systems},
  volume={24},
  number={1},
  pages={178--190},
  year={2022},
  publisher={IEEE}
}

@article{wang2021gan,
  title={GAN and multi-agent DRL based decentralized traffic light signal control},
  author={Wang, Zixin and Zhu, Hanyu and He, Mingcheng and Zhou, Yong and Luo, Xiliang and Zhang, Ning},
  journal={IEEE Transactions on Vehicular Technology},
  volume={71},
  number={2},
  pages={1333--1348},
  year={2021},
  publisher={IEEE}
}

@article{hu2024multi,
  title={A multi-agent deep reinforcement learning approach for traffic signal coordination},
  author={Hu, Ta-Yin and Li, Zhuo-Yu},
  journal={IET Intelligent Transport Systems},
  volume={18},
  number={8},
  pages={1428--1444},
  year={2024},
  publisher={Wiley Online Library}
}

@article{wang2020large,
  title={Large-scale traffic signal control using a novel multiagent reinforcement learning},
  author={Wang, Xiaoqiang and Ke, Liangjun and Qiao, Zhimin and Chai, Xinghua},
  journal={IEEE Transactions on Cybernetics},
  volume={51},
  number={1},
  pages={174--187},
  year={2020},
  publisher={IEEE}
}

@inproceedings{tan2020multi,
  title={Multi-agent bootstrapped deep q-network for large-scale traffic signal control},
  author={Tan, Tian and Chu, Tianshu and Wang, Jie},
  booktitle={2020 IEEE Conference on Control Technology and Applications (CCTA)},
  pages={358--365},
  year={2020},
  organization={IEEE}
}

@article{wu2020multi1,
  title={Multi-agent deep reinforcement learning for urban traffic light control in vehicular networks},
  author={Wu, Tong and Zhou, Pan and Liu, Kai and Yuan, Yali and Wang, Xiumin and Huang, Huawei and Wu, Dapeng Oliver},
  journal={IEEE Transactions on Vehicular Technology},
  volume={69},
  number={8},
  pages={8243--8256},
  year={2020},
  publisher={IEEE}
}

@article{bokade2023multi,
  title={Multi-agent reinforcement learning based on representational communication for large-scale traffic signal control},
  author={Bokade, Rohit and Jin, Xiaoning and Amato, Christopher},
  journal={IEEE Access},
  volume={11},
  pages={47646--47658},
  year={2023},
  publisher={IEEE}
}

@inproceedings{wu2020multi2,
  title={Multi-agent reinforcement learning for traffic signal control: Algorithms and robustness analysis},
  author={Wu, Chunliang and Ma, Zhenliang and Kim, Inhi},
  booktitle={2020 IEEE 23rd International Conference on Intelligent Transportation Systems (ITSC)},
  pages={1--7},
  year={2020},
  organization={IEEE}
}

@article{ge2021multi,
  title={Multi-agent transfer reinforcement learning with multi-view encoder for adaptive traffic signal control},
  author={Ge, Hongwei and Gao, Dongwan and Sun, Liang and Hou, Yaqing and Yu, Chao and Wang, Yuxin and Tan, Guozhen},
  journal={IEEE Transactions on Intelligent Transportation Systems},
  volume={23},
  number={8},
  pages={12572--12587},
  year={2021},
  publisher={IEEE}
}

@article{zhang2022neighborhood,
  title={Neighborhood cooperative multiagent reinforcement learning for adaptive traffic signal control in epidemic regions},
  author={Zhang, Chengwei and Tian, Yu and Zhang, Zhibin and Xue, Wanli and Xie, Xiaofei and Yang, Tianpei and Ge, Xin and Chen, Rong},
  journal={IEEE Transactions on Intelligent Transportation Systems},
  volume={23},
  number={12},
  pages={25157--25168},
  year={2022},
  publisher={IEEE}
}

@article{lee2019reinforcement,
  title={Reinforcement learning for joint control of traffic signals in a transportation network},
  author={Lee, Jincheol and Chung, Jiyong and Sohn, Keemin},
  journal={IEEE Transactions on Vehicular Technology},
  volume={69},
  number={2},
  pages={1375--1387},
  year={2019},
  publisher={IEEE}
}

@article{luo2024reinforcement,
  title={Reinforcement learning for traffic signal control in hybrid action space},
  author={Luo, Haoqing and Bie, Yiming and Jin, Sheng},
  journal={IEEE Transactions on Intelligent Transportation Systems},
  year={2024},
  publisher={IEEE}
}

@article{liu2023traffic,
  title={A traffic light control method based on multi-agent deep reinforcement learning algorithm},
  author={Liu, Dongjiang and Li, Leixiao},
  journal={Scientific Reports},
  volume={13},
  number={1},
  pages={9396},
  year={2023},
  publisher={Nature Publishing Group UK London}
}

@article{wang2020stmarl,
  title={STMARL: A spatio-temporal multi-agent reinforcement learning approach for cooperative traffic light control},
  author={Wang, Yanan and Xu, Tong and Niu, Xin and Tan, Chang and Chen, Enhong and Xiong, Hui},
  journal={IEEE Transactions on Mobile Computing},
  volume={21},
  number={6},
  pages={2228--2242},
  year={2020},
  publisher={IEEE}
}

@article{wang2021adaptive,
  title={Adaptive Traffic Signal Control for large-scale scenario with Cooperative Group-based Multi-agent reinforcement learning},
  author={Wang, Tong and Cao, Jiahua and Hussain, Azhar},
  journal={Transportation Research Part C: Emerging Technologies},
  volume={125},
  pages={103046},
  year={2021},
  publisher={Elsevier}
}

@article{li2021network,
  title={Network-wide traffic signal control optimization using a multi-agent deep reinforcement learning},
  author={Li, Zhenning and Yu, Hao and Zhang, Guohui and Dong, Shangjia and Xu, Cheng-Zhong},
  journal={Transportation Research Part C: Emerging Technologies},
  volume={125},
  pages={103059},
  year={2021},
  publisher={Elsevier}
}

@article{mckenzie2022modern,
  title={Modern value based reinforcement learning: A chronological review},
  author={McKenzie, Mark C and McDonnell, Mark D},
  journal={IEEE Access},
  volume={10},
  pages={134704--134725},
  year={2022},
  publisher={IEEE}
}

@article{li2017deep,
  title={Deep reinforcement learning: An overview},
  author={Li, Yuxi},
  journal={arXiv preprint arXiv:1701.07274},
  year={2017}
}

@article{farazi2021deep,
  title={Deep reinforcement learning in transportation research: A review},
  author={Farazi, Nahid Parvez and Zou, Bo and Ahamed, Tanvir and Barua, Limon},
  journal={Transportation Research Interdisciplinary Perspectives},
  volume={11},
  pages={100425},
  year={2021},
  publisher={Elsevier}
}

@article{gong2021research,
  title={Research review for broad learning system: Algorithms, theory, and applications},
  author={Gong, Xinrong and Zhang, Tong and Chen, CL Philip and Liu, Zhulin},
  journal={IEEE Transactions on Cybernetics},
  volume={52},
  number={9},
  pages={8922--8950},
  year={2021},
  publisher={IEEE}
}

@article{jiang2023pa,
  title={Pa-count: passenger counting in vehicles using wi-fi signals},
  author={Jiang, Hongbo and Chen, Siyu and Xiao, Zhu and Hu, Jingyang and Liu, Jiangchuan and Dustdar, Schahram},
  journal={IEEE Transactions on Mobile Computing},
  volume={23},
  number={4},
  pages={2684--2697},
  year={2023},
  publisher={IEEE}
}

@article{papakis2021convolutional,
  title={Convolutional neural network-based in-vehicle occupant detection and classification method using second strategic highway research program cabin images},
  author={Papakis, Ioannis and Sarkar, Abhijit and Svetovidov, Andrei and Hickman, Jeffrey S and Abbott, A Lynn},
  journal={Transportation Research Record},
  volume={2675},
  number={8},
  pages={443--457},
  year={2021},
  publisher={SAGE Publications Sage CA: Los Angeles, CA}
}

@article{gong2022using,
  title={Using machine learning to estimate pedestrian and bicyclist count of intersection by Bluetooth low energy},
  author={Gong, Yaobang and Abdel-Aty, Mohamed},
  journal={Journal of Transportation Engineering, Part A: Systems},
  volume={148},
  number={1},
  pages={04021101},
  year={2022},
  publisher={American Society of Civil Engineers}
}

@article{roncoli2023estimating,
  title={Estimating on-board passenger comfort in public transport vehicles using incomplete automatic passenger counting data},
  author={Roncoli, Claudio and Chandakas, Ektoras and Kaparias, Ioannis},
  journal={Transportation Research Part C: Emerging Technologies},
  volume={146},
  pages={103963},
  year={2023},
  publisher={Elsevier}
}

@article{jia2025multi,
  title={Multi-Agent Deep Reinforcement Learning for Large-Scale Traffic Signal Control with Spatio-Temporal Attention Mechanism},
  author={Jia, Wenzhe and Ji, Mingyu},
  journal={Applied Sciences},
  volume={15},
  number={15},
  pages={8605},
  year={2025},
  publisher={MDPI}
}

@article{fereidooni2025multi,
  title={Multi Agent Optimizing Traffic Light Signals Using Deep Reinforcement Learning},
  author={Fereidooni, Zahra and Palesi, LA Ipsaro and Nesi, Paolo},
  journal={IEEE Access},
  year={2025},
  publisher={IEEE}
}

@article{shen2025hierarchical,
  title={Hierarchical reinforcement learning-based traffic signal control},
  author={Shen, Jiajing},
  journal={Scientific Reports},
  volume={15},
  number={1},
  pages={32862},
  year={2025},
  publisher={Nature Publishing Group UK London}
}

@article{liu2025globallight,
  title={GlobalLight: Exploring global influence in multi-agent deep reinforcement learning for large-scale traffic signal control},
  author={Liu, Yilin and Liang, Jintao and Zhang, Yifeng and Gong, Ping and Luo, Guiyang and Yuan, Quan and Li, Jinglin},
  journal={Neurocomputing},
  volume={637},
  pages={130065},
  year={2025},
  publisher={Elsevier}
}

@inproceedings{lai2025llmlight,
  title={Llmlight: Large language models as traffic signal control agents},
  author={Lai, Siqi and Xu, Zhao and Zhang, Weijia and Liu, Hao and Xiong, Hui},
  booktitle={Proceedings of the 31st ACM SIGKDD Conference on Knowledge Discovery and Data Mining},
  pages={2335--2346},
  year={2025}
}

@article{satheesh2025constrained,
  title={A constrained multi-agent reinforcement learning approach to autonomous traffic signal control},
  author={Satheesh, Anirudh and Powell, Keenan},
  journal={Journal on Autonomous Transportation Systems},
  year={2025},
  publisher={ACM New York, NY}
}

@article{yoon2025decentralized,
  title={Decentralized and Communication-Based Multi-Agent Traffic Signal Control Model Employing a Graph Representation for the State},
  author={Yoon, Jinwon and Ahn, Kyuree and Lee, Kanghoon and Park, Jinkyoo and Yeo, Hwasoo},
  journal={IEEE Transactions on Intelligent Transportation Systems},
  year={2025},
  publisher={IEEE}
}

@article{el2013multiagent,
  title={Multiagent reinforcement learning for integrated network of adaptive traffic signal controllers (MARLIN-ATSC): methodology and large-scale application on downtown Toronto},
  author={El-Tantawy, Samah and Abdulhai, Baher and Abdelgawad, Hossam},
  journal={IEEE Transactions on Intelligent Transportation Systems},
  volume={14},
  number={3},
  pages={1140--1150},
  year={2013},
  publisher={IEEE}
}

@inproceedings{chu2016large,
  title={Large-scale traffic grid signal control with regional reinforcement learning},
  author={Chu, Tianshu and Qu, Shuhui and Wang, Jie},
  booktitle={2016 American Control Conference (acc)},
  pages={815--820},
  year={2016},
  organization={IEEE}
}

@inproceedings{ma2020feudal,
  title={Feudal multi-agent deep reinforcement learning for traffic signal control},
  author={Ma, Jinming and Wu, Feng},
  booktitle={Proceedings of the 19th International Conference on Autonomous Agents and Multiagent Systems},
  pages={816--824},
  year={2020}
}

@inproceedings{tao2023network,
  title={Network Clustering-Based Multi-Agent Reinforcement Learning for Large-Scale Traffic Signal Control},
  author={Tao, Zhicheng and Li, Chao and Yang, Qinmin},
  booktitle={2023 International Annual Conference on Complex Systems and Intelligent Science (CSIS-IAC)},
  pages={326--331},
  year={2023},
  organization={IEEE}
}

@article{wang2025towards,
  title={Towards multi-agent reinforcement learning based traffic signal control through spatio-temporal hypergraphs},
  author={Wang, Kang and Shen, Zhishu and Lei, Zhen and Liu, Xianhui and Zhang, Tiehua},
  journal={IEEE Transactions on Mobile Computing},
  year={2025},
  publisher={IEEE}
}

@article{zhang2026human,
  title={Human-centric traffic signal control for equity: A multi-agent action branching deep reinforcement learning approach},
  author={Zhang, Xiaocai and Nassir, Neema and Chan, Lok Sang and Haghani, Milad},
  journal={Engineering Applications of Artificial Intelligence},
  volume={177},
  number={1},
  pages={114904},
  year={2026}
}

@inproceedings{zhang2026realistic,
  title={Realistic Curriculum Reinforcement Learning for Autonomous and Sustainable Marine Vessel Navigation},
  author={Zhang, Xiaocai and Xiao, Zhe and Liang, Maohan and Liu, Tao and Li, Haijiang and Zhang, Wenbin},
  booktitle={Proceedings of the AAAI Conference on Artificial Intelligence},
  volume={40},
  number={46},
  pages={39612--39619},
  year={2026}
}

@article{ramadhan2020application,
  title={Application of area traffic control using the max-pressure algorithm},
  author={Ramadhan, Satria A and Sutarto, Herman Y and Kuswana, Gilang S and Joelianto, Endra},
  journal={Transportation Planning and Technology},
  volume={43},
  number={8},
  pages={783--802},
  year={2020},
  publisher={Taylor \& Francis}
}

@article{chan2026multi,
  title={Multi-Objective Multi-Step Adaptive Traffic Control (MOMSATC): Prioritising Pedestrians for a Safe and Sustainable Transport Development},
  author={Chan, Lok Sang and Zhang, Xiaocai and Nassir, Neema and Sarvi, Majid},
  journal={IET Intelligent Transport Systems},
  volume={20},
  number={1},
  pages={e70150},
  year={2026},
  publisher={Wiley Online Library}
}

@article{Chan2026regulating,
  title={Regulating jaywalking behaviour in adaptive traffic signal control using a novel deep reinforcement learning approach},
  author={Chan, Lok Sang and Zhang, Xiaocai and Nassir, Neema and Sarvi, Majid},
  journal={Multimodal Transportation},
  volume={5},
  number={2},
  pages={100271},
  year={2026},
  publisher={Elsevier}
}

@article{duan2025bayesian,
  title={Bayesian Critique-Tune-Based Reinforcement Learning With Adaptive Pressure for Multi-Intersection Traffic Signal Control},
  author={Duan, Wenchang and Gao, Zhenguo and He, Jiwan and Xian, Jinguo},
  journal={IEEE Transactions on Intelligent Transportation Systems},
  year={2025},
  publisher={IEEE}
}

@article{xu2024graph,
  title={A graph deep reinforcement learning traffic signal control for multiple intersections considering missing data},
  author={Xu, Dongwei and Yu, Zefeng and Liao, Xiangwang and Guo, Haifeng},
  journal={IEEE Transactions on Vehicular Technology},
  volume={73},
  number={12},
  pages={18307--18319},
  year={2024},
  publisher={IEEE}
}

\end{document}